\documentclass{article}

\PassOptionsToPackage{square,numbers,sort&compress}{natbib}

\usepackage[preprint]{neurips_2025}

\usepackage{graphicx}
\usepackage{wrapfig}
\usepackage{makecell}
\usepackage{subcaption}
\usepackage{tabularx}
\usepackage{colortbl}

\usepackage[utf8]{inputenc} % allow utf-8 input
\usepackage[T1]{fontenc} % use 8-bit T1 fonts
\usepackage{hyperref} % hyperlinks
\usepackage{url} % simple URL typesetting
\usepackage{booktabs} % professional-quality tables
\usepackage{amsfonts} % blackboard math symbols
\usepackage{nicefrac} % compact symbols for 1/2, etc.
\usepackage{microtype} % microtypography
\usepackage{xcolor} % colors
\usepackage{pifont}

\usepackage{amsmath}
\usepackage{amssymb}
\usepackage{mathtools}
\usepackage{amsthm}

\usepackage{adjustbox}
\usepackage{multirow}
\usepackage{enumitem}

\definecolor{cxcolor}{HTML}{00675B}

\newcommand{\cmark}{{\color[rgb]{0.6,0.8,0.5}\ding{51}}}
\newcommand{\xmark}{{\color{red}\ding{55}}}

\title{FLAME-MoE: A Transparent End-to-End Research Platform for Mixture-of-Experts Language Models}

\author{%
  Hao Kang$^{2}$\thanks{Equal contribution. Author order determined alphabetically.} \quad Zichun Yu$^{1}$\footnotemark[1] \quad Chenyan Xiong$^{1,3}$ \\
  $^{1}$Language Technologies Institute \\
  $^{2}$School of Computer Science \\
  $^{3}$Foundation and Language Model Center \\
  Carnegie Mellon University \\
  \texttt{\{haok, zichunyu, cx\}@andrew.cmu.edu}
}

\begin{document}
    \maketitle
    % \begin{abstract}
% Recent large language models such as Gemini-1.5, DeepSeek-V3 and Llama-4 increasingly adopt Mixture-of-Experts (MoE) architectures, which activate only a subset of parameters per token and thereby offer substantial computational savings. Yet academic researchers still lack a fully open, end-to-end MoE platform for investigating scaling, routing, and expert behavior. We release FLAME-MoE, a completely open-source research suite composed of seven decoder-only models, ranging from 38M to 1.7B active parameters, whose architecture choices—64 experts with top-8 gating and 2 shared experts—mirror mainstream production LLMs. All training data pipelines, scripts, logs, and checkpoints are publicly available to enable reproducible experimentation. Across six evaluation tasks, FLAME-MoE improves average accuracy by up to 3.4 points over dense baselines trained with identical FLOPs. Leveraging the fully transparent training trace, we conduct comprehensive analyses and observe that (i) experts progressively focus on distinct token subsets over time, (ii) co-activation matrices remain sparse throughout training which reflects diverse expert usage, and (iii) routing decisions are learned swiftly in the initial training phase. We open-source all code, training logs, and model checkpoints at \url{https://github.com/cmu-flame/FLAME-MoE}.
% \end{abstract}

\begin{abstract}
Recent large language models such as Gemini-1.5, DeepSeek-V3, and Llama-4 increasingly adopt Mixture-of-Experts (MoE) architectures, which offer strong efficiency-performance trade-offs by activating only a fraction of the model per token. Yet academic researchers still lack a fully open, end-to-end MoE platform for investigating scaling, routing, and expert behavior. We release FLAME-MoE, a completely open-source research suite composed of seven decoder-only models, ranging from 38M to 1.7B active parameters, whose architecture—64 experts with top-8 gating and 2 shared experts—closely reflects modern production LLMs. All training data pipelines, scripts, logs, and checkpoints are publicly available to enable reproducible experimentation. Across six evaluation tasks, FLAME-MoE improves average accuracy by up to 3.4 points over dense baselines trained with identical FLOPs. Leveraging full training trace transparency, we present initial analyses showing that (i) experts increasingly specialize on distinct token subsets, (ii) co-activation matrices remain sparse, reflecting diverse expert usage, and (iii) routing behavior stabilizes early in training. All code, training logs, and model checkpoints are available at \url{https://github.com/cmu-flame/FLAME-MoE}.
\end{abstract}

    \section{Introduction}

Mixture-of-Experts (MoE) architectures have emerged as a powerful paradigm for scaling large language models efficiently. By activating only a sparse subset of expert modules per input token, MoE models significantly expand the number of trainable parameters without a commensurate increase in training or inference cost. This approach enables high effective capacity with strong computational efficiency, and has been adopted in a range of state-of-the-art systems—both open and closed-source—including Gemini-1.5, DeepSeek-V3, and Llama-4~\cite{team2024gemini, liu2024deepseekv3, meta2025llama4}. As these architectures gain traction, there is a growing need for transparent, reproducible frameworks to facilitate systematic study of MoE behavior, limitations, and scaling dynamics.

Recent open-source efforts such as OLMoE and OpenMoE have made notable progress toward building accessible and performant MoE models, releasing architectural variants, pretrained checkpoints, and analyses of routing and expert specialization~\cite{muennighoff2024olmoe, xue2024openmoe}. These contributions represent valuable steps in expanding the empirical foundation for MoE research. However, they primarily emphasize architectural design and downstream performance, offering limited support for systematic, end-to-end investigation of MoE training dynamics, scaling behavior, and routing evolution. As a result, unlike the dense model community—which has benefited from comprehensive frameworks such as Pythia~\cite{biderman2023pythia}—the MoE research landscape has yet to converge around a shared experimental platform that enables reproducible and extensible analysis across model scales.

To address the gap in accessible and reproducible research on MoE language models, we introduce FLAME-MoE: a transparent, robust platform for controlled experimentation across model scales, sparsity levels, and architectures. It includes seven decoder-only MoE models (38M–1.7B active parameters), each with 64 experts per layer, top-8 gating, and two shared experts, following DeepSeek-V2~\cite{liu2024deepseekv2} and OLMoE~\cite{muennighoff2024olmoe}. Using empirical scaling laws, we allocate compute-optimal training budgets to ensure fair, efficient pretraining. As summarized in Table~\ref{tab:oss}, FLAME-MoE is the only MoE platform offering full openness—code, data, checkpoints, routing logs, and evaluation results—across a wide range of model sizes, supporting reproducibility, accessibility, and cross-scale analysis.

Empirical evaluations on 6 downstream tasks show that FLAME-MoE consistently outperforms dense counterparts trained under equal compute, with accuracy gains ranging from 1.8 to 3.4 points. More importantly, our training traces reveal nuanced dynamics of MoE models: expert specialization emerges early and intensifies over time; expert co-activation remains sparse and stable; and routing behaviors quickly converge during early pretraining. These findings underscore FLAME-MoE's value as a tool for in-depth analysis of expert dynamics, sparsity effects, and routing evolution.

FLAME-MoE is not merely a model release, but a comprehensive platform for advancing research on sparse language models. Our main contributions include: (1) a fully transparent infrastructure for end-to-end MoE experimentation, including models, training scripts, routing traces, and evaluation pipelines; (2) a suite of consistent, modular architectures spanning a wide range of active parameters to facilitate rigorous comparison across scales and with dense baselines; and (3) analytical capabilities that enable fine-grained investigation of routing behavior, expert specialization, load balancing, and parallelization strategies—questions that remain difficult to address without full access to the training process. By opening every stage of model development to the research community, we hope FLAME-MoE could establish a principled and extensible foundation for future work on Mixture-of-Experts.

% \begin{minipage}{1.0\linewidth}
\begin{table}
    \centering
    \caption{Openness comparison between open-source MoE models.}
    
    \resizebox{0.8\textwidth}{!}{%
    \begin{tabular}{lcccccr}
    \toprule
    \textbf{Model} & \textbf{Code} & \textbf{Data} & \textbf{All Ckpts} & \textbf{Logs} & \textbf{\#Scales} & \textbf{Active/Total Parameters} \\
    \midrule
    JetMoE~\cite{shen2024jetmoe} & \cmark & \xmark & \xmark & \xmark & 1 & 2B/8B \\
    OpenMoE~\cite{xue2024openmoe} & \cmark & \cmark & \xmark & \xmark & 3 & 339M/650M-6.8B/34B \\
    OLMoE~\cite{muennighoff2024olmoe} & \cmark & \cmark & \cmark & \cmark & 1 & 1.3B/6.9B \\
    FLAME-MoE & \cmark & \cmark & \cmark & \cmark & 7 & 38M/100M-1.7B/10.3B \\
    \bottomrule
    \end{tabular}
    }
    \label{tab:oss}
\end{table}
% \vspace{-0.5cm}
% \end{minipage}

    \section{Related Work}

Mixture-of-Experts models aim to increase model capacity without a commensurate rise in computational cost by selectively activating a sparse subset of expert modules per input~\cite{shazeer2017outrageously,jiang2024mixtral}. Foundational works such as GShard~\cite{lepikhin2020gshard} and Switch Transformer~\cite{fedus2022switch} demonstrated the efficacy of sparse expert routing in enabling the training of large-scale models with substantially reduced floating point operations (FLOPs), while maintaining competitive performance. These models introduced top-$k$ routing mechanisms alongside auxiliary losses to ensure balanced expert utilization, thus establishing key design principles for scalable sparse training. Subsequent developments, including BASE Layers~\cite{lewis2021base}, extended this paradigm by employing global routing via optimal transport to enhance load balancing and training stability. Later models such as DeepSeek-V3~\cite{liu2024deepseekv3} and M6-T~\cite{yang2021m6} scaled these approaches further, incorporating mechanisms such as shared experts and expert prototyping to improve convergence and efficiency, particularly in multilingual and long-context scenarios.

Recent work has also examined the robustness and specialization of expert routing in MoE architectures~\cite{chi2022representation,xue2024openmoe,muennighoff2024olmoe}. For instance, \citet{chi2022representation} address the issue of representation collapse—where token embeddings become overly concentrated around expert centroids—by projecting tokens and expert keys onto a hypersphere prior to computing routing scores. This modification enhances assignment diversity and yields improved performance on multilingual benchmarks. Other studies have investigated the evolution of routing dynamics throughout training~\cite{xue2024openmoe,muennighoff2024olmoe}, noting that expert activation patterns frequently correlate with input position and frequency. Collectively, these findings underscore the importance of not only architectural efficiency but also the stability of sparse computation.

In parallel, large-scale dense model platforms have played a critical role in advancing empirical research by enhancing reproducibility and transparency. Pythia~\cite{biderman2023pythia}, for example, comprises a suite of 16 decoder-only models trained on a common dataset, with full checkpoint availability and precise training reconstructions. This design enables fine-grained analyses of scaling laws, memorization behaviors, and training dynamics. Similarly, OPT~\cite{zhang2022opt} released dense models ranging from 125M to 175B parameters, along with training code, hyperparameters, and logs, setting a precedent for open large-scale experimentation. Pythia has served as the foundation for a range of downstream studies and systems, including instruction-tuned models such as Dolly 2.0~\cite{dolly2023}. Researchers have leveraged the Pythia suite to examine memorization~\cite{huang2024demystifying,biderman2023emergent}, convergence patterns~\cite{martinez2024tending}, and internal circuit development~\cite{tigges2024llm}. Its consistent data ordering across model sizes enables rigorous scaling analyses, supporting investigations into privacy risks~\cite{akkus2024generated}, interpretability~\cite{skean2025layer}, and alignment techniques such as Direct Preference Optimization (DPO)~\cite{o2024attributing}.

While early MoE systems were predominantly developed on proprietary infrastructure, recent open-source initiatives have significantly broadened accessibility. Notable efforts such as OpenMoE~\cite{xue2024openmoe} and OLMoE~\cite{muennighoff2024olmoe} have released models, training frameworks, and diagnostic tools, facilitating the study of expert specialization and routing behavior while narrowing the reproducibility gap. From a systems perspective, multiple frameworks have emerged to support efficient MoE training and deployment. DeepSpeed-MoE~\cite{rajbhandari2022deepspeed} and Tutel~\cite{hwang2023tutel} offer optimized all-to-all communication, expert parallelism, and high-throughput inference, supporting models at the trillion-parameter scale. FastMoE~\cite{he2021fastmoe} provides a lightweight PyTorch extension with minimal integration overhead, while Megatron-LM~\cite{shoeybi2019megatron} integrates MoE support alongside tensor, expert, and pipeline parallelism. These systems collectively enable scalable, production-grade sparse modeling via optimized infrastructure.

    \section{Model Architecture}
\label{sec:architecture}

Our model architecture builds upon recent advancements in MoE models, particularly drawing inspirations from DeepSeek-V2~\cite{liu2024deepseekv2} and OLMoE~\cite{muennighoff2024olmoe}. 
Specifically, we adopt a decoder-only transformer architecture consisting of $N_L$ layers, where all feedforward network (FFN) sublayers except for the first one are replaced by an MoE layer.

\subsection{MoE Layer}
Each MoE layer comprises $N_E$ expert networks (FFNs) and a router mechanism. For an input token representation $x$, we employ Top-$k$ routing, where only the $k$ experts with the highest routing scores are selected to process the token:
\begin{equation}
    \text{MoE}(x) = \sum_{i \in \text{Top-}k(r(x))} \text{softmax}(r(x))_i E_i(x)
    \label{eq:moe_module}
\end{equation}
where $E_i$ denotes the $i$-th expert network and $r$ is a router network that
computes router scores $r(x)$ for each of the $N_E$ experts based on the input token representation $x$. 
The $\text{Top-}k$ function selects the indices of the $k$ experts yielding the highest scores. 
These scores are normalized via a softmax function, resulting in gating weights $\text{softmax}(r(x))_i$. 
The final output of the MoE layer is a weighted sum: the output $E_i(x)$ of each selected expert $i \in \text{Top-}k(r(x))$ is multiplied by its corresponding gating weight, and these products are summed together.

Following best practice and findings from OLMoE and DeepSeek-V2, we set the total number of experts $N_E = 64$ per MoE layer and activate $k=8$ experts per token. Adapting the strategy from DeepSeek-V2, 2 of these 8 active experts are designated as shared experts, meaning they are activated for every token, providing a baseline computation path. The remaining 6 experts are routed experts, 
selected dynamically by the router based on the input token. 

\subsection{Training Loss}
Training MoE models effectively often requires auxiliary loss functions to encourage a balanced load across experts and prevent routing collapse. 
% We adopt an expert-level load balancing loss, similar to those used in OLMoE and DeepSeek-V2:
% \begin{align}
%     \mathcal{L}_{\text{ExpBal}} &= \alpha_1 \sum_{i=1}^{N_r} f_i P_i, \label{eq:expert_balance} \\
%     \text{where } f_i = \frac{N_r}{K_r T} \sum_{t=1}^{T} &\mathbf{1}(\text{Token } t \text{ selects Expert } i)~\text{and } P_i = \frac{1}{T} \sum_{t=1}^{T} s_{i,t}.
% \end{align}
% Here, $f_i$ represents the fraction of tokens dispatched to expert $i$ within a sequence (or batch), scaled by $N_r/K_r$. $P_i$ is the average routing probability (or score $s_{i,t}$) assigned to expert $i$ by the router over the $T$ tokens in the sequence. $N_r$ is the number of routed experts, $\mathbf{1}(\cdot)$ denotes the indicator function, 
% $K_r$ is a normalization constant and $\alpha_1$ is a hyperparameter controlling the loss weight.
First, we adopt a load balancing loss from~\citet{shazeer2017outrageously} to ensure that tokens are distributed as evenly as possible among the $N_E$ experts, which improves both model utilization and parallelism. 
For each routing operation, given $N_E$ experts and a batch of $T$ tokens, 
the load balancing loss $\mathcal{L}_{\text{LB}}$ is defined as:
\begin{align}
    \mathcal{L}_{\text{LB}} &= N_E \cdot \sum_{i=1}^{N_E} m_i \cdot P_i,
    \label{eq:load_balance} \\
    \text{where } m_i = \frac{1}{T} \sum_{j=1}^T &\mathbf{1}(x_j)_i \text{ and } P_i = \frac{1}{T} \sum_{j=1}^T \text{softmax}(r(x_j))_i.
\end{align}
Here, $m_i$ is the fraction of tokens dispatched to expert $i$;
$P_i$ is the average gating weight assigned to expert $i$ by the router over the batch;
$\mathbf{1}(x_j)_i$ is an indicator function that equals 1 if token $x_j$ is routed to expert $i$, and 0 otherwise. 
This loss encourages the router to assign tokens more uniformly across all experts, mitigating the risk of expert underutilization and routing collapse.

In addition to the load balancing loss, we incorporate a router z-loss~\cite{zoph2022st} to further stabilize MoE training. 
The z-loss penalizes large-magnitude router logits, encouraging the router to produce outputs with smaller absolute values, 
which helps reduce numerical instability and round-off errors when passing through the gating function. 
Formally, the router z-loss is defined as:
\begin{equation}
    \mathcal{L}_{\text{RZ}}(x) = \frac{1}{T} \sum_{i=1}^T \left( \log \sum_{j=1}^{N_E} \exp(r(x_i)_j) \right)^2,
    \label{eq:z_loss}
\end{equation}
where $r(x_i)_j$ denotes the router logit for token $x_i$ and expert $j$. 
This loss encourages the router logits to remain well-behaved, improving the stability and efficacy of MoE training.

The final training objective combines the standard cross-entropy loss ($\mathcal{L}_{\text{CE}}$) with two auxiliary losses:
\begin{equation}
    \mathcal{L}_{\text{train}} = \mathcal{L}_{\text{CE}} + \gamma \mathcal{L}_{\text{LB}} + \eta \mathcal{L}_{\text{RZ}},
    \label{eq:total_loss}
\end{equation}
where $\gamma$ and $\eta$ are two hyperparameters controlling the strengths of the load balancing loss and the router z-loss. 
Following OLMoE, we set $\gamma = 0.01$ and $\eta = 0.001$ for all experiments.
    \section{Scaling Law Study for FLAME-MoE}
\label{sec:scaling}

\begin{figure*}
    \centering
    \vspace{-0.3cm}
    \begin{subfigure}
        {0.23\textwidth}
        \centering
        \includegraphics[width=1.0\linewidth]{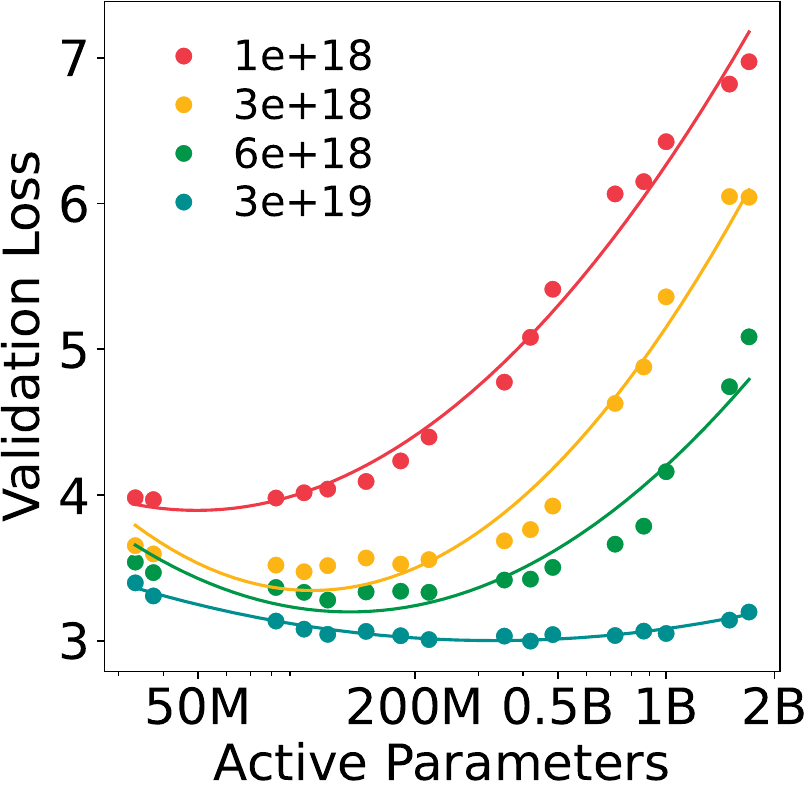}
        \caption{IsoFLOP profiles.}
        \label{fig:scaling-isoflop}
    \end{subfigure}
    ~
    \begin{subfigure}
        {0.23\textwidth}
        \centering
        \includegraphics[width=1.0\linewidth]{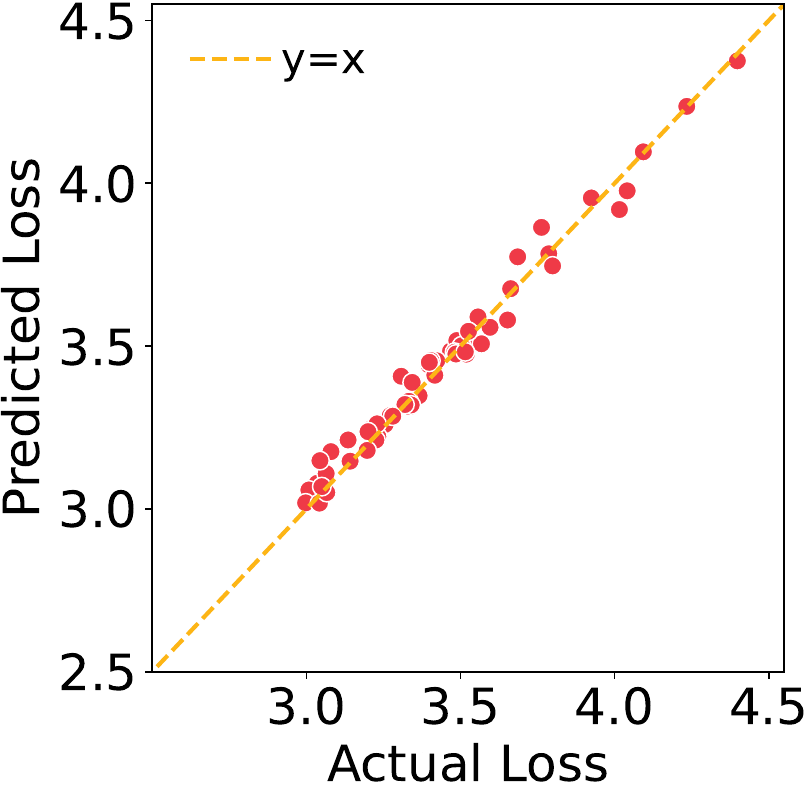}
        \caption{Parametric loss.}
        \label{fig:scaling-parametric}
    \end{subfigure}
    ~
    \begin{subfigure}
        {0.258\textwidth}
        \centering
        \includegraphics[width=1.0\linewidth]{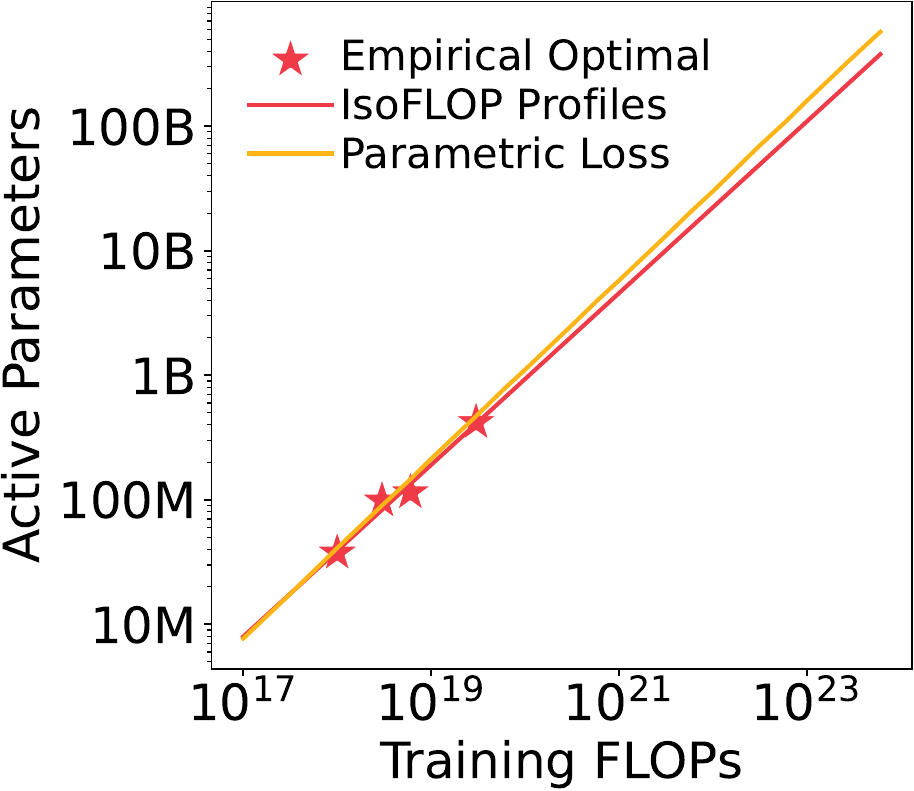}
        \caption{Fitted scaling law.}
        \label{fig:scaling-fitted}
    \end{subfigure}
    ~
    \begin{subfigure}
        {0.23\textwidth}
        \centering
        \includegraphics[width=1.0\linewidth]{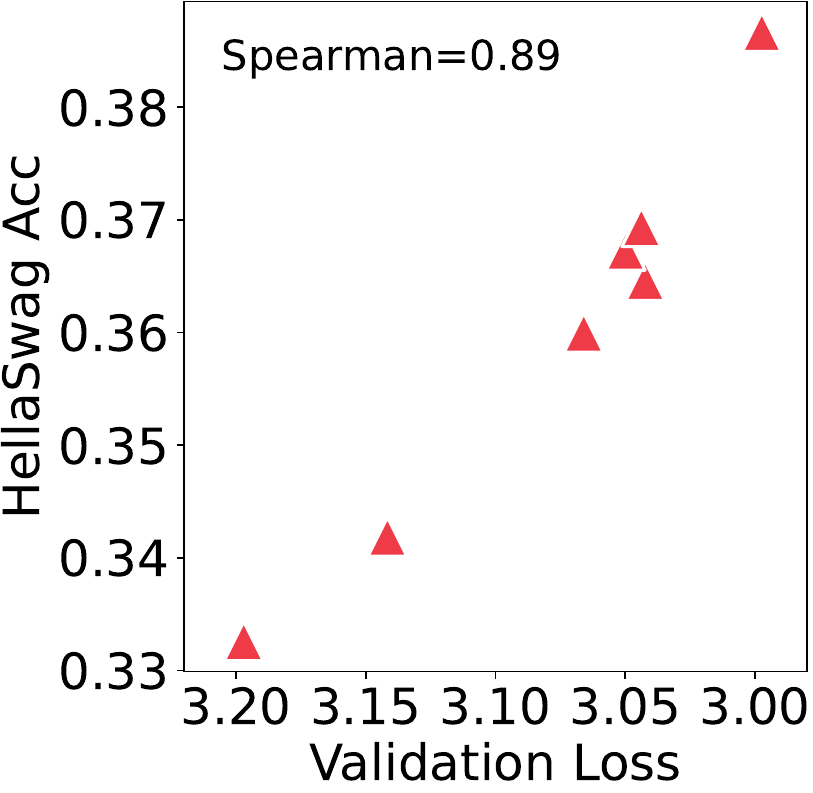}
        \caption{Generalization ability.}
        \label{fig:scaling-generalization}
    \end{subfigure}
    \caption{Scaling law experiments: (a) IsoFLOP profiles; (b) parametric loss function fitting; (c) fitted scaling law; (d) generalization from validation loss to downstream performance.}
    \label{fig:scaling}
    % \vspace{-0.5cm}
\end{figure*}

\begin{table}
    \small
    \centering
    \caption{Model cards of 7 FLAME-MoE models.}
    % For all models, we set the total number of experts as 64 and activate 8 experts per token, where 2 of 8 are shared experts.
    \resizebox{1.0\textwidth}{!}{%
    \begin{tabular}{rrrrrrrrr}
    \toprule
    \textbf{Active-Total} & \textbf{Layers} & \textbf{Hidden Size} & \textbf{FFN Hidden} & \textbf{MoE FFN Hidden} & \textbf{Total/Active/Shared} & \textbf{\#FLOPs} & \textbf{H100 Hours} & \textbf{Tokens} \\
    \midrule
    38M-100M & 9 & 256 & 1368 & 176 & 64/8/2 & 1.0e18 & 5.7 & 4.4B \\
    98M-349M & 9 & 512 & 2736 & 352 & 64/8/2 & 3.0e18 & 8.2 & 5.0B \\
    115M-459M & 12 & 512 & 2736 & 352 & 64/8/2 & 6.0e18 & 20.9 & 8.7B \\
    290M-1.3B & 9 & 1024 & 5472 & 704 & 64/8/2 & 2.0e19 & 63.4 & 11.4B \\
    419M-2.2B & 15 & 1024 & 5472 & 704 & 64/8/2 & 3.0e19 & 66.4 & 11.9B \\
    721M-3.8B & 12 & 1536 & 8208 & 1056 & 64/8/2 & 8.0e19 & 172.8 & 18.4B \\
    1.7B-10.3B & 18 & 2048 & 10944 & 1408 & 64/8/2 & 2.4e20 & 560.5 & 23.1B \\
    \bottomrule
    \end{tabular}
    }
    \label{tab:config}
\end{table}

Determining the optimal allocation between the model size and the number of training tokens for a fixed compute budget (FLOPs) is crucial to training LLMs efficiently. 
% In FLAME-MoE architecture, we follow DeepSeek-V2~\cite{liu2024deepseekv2} to fix the total number of experts and the number of active experts per token. 
% Consequently, the number of active parameters uniquely determines the model configuration. 
% assuming that other architectural factors such as hidden size and number of layers are set according to best practices established for dense models~\cite{biderman2023pythia}.
As we fix the total number of experts and the number of active experts per token for FLAME-MoE, 
we concentrate on the number of active parameters as the primary variable for our scaling law study.
This simplification allows us to adapt established methodologies in dense models for finding compute-optimal MoE configurations.

Specifically, we investigate two primary approaches employed in Chinchilla~\cite{Chinchilla}, to determine the optimal $N_{\text{active}}$ for our MoE models given a fixed FLOPs budget, 
IsoFLOP profiles (§\ref{sec:isoflop}) and parametric loss function fitting (§\ref{sec:parametric_loss}). Finally, we apply our fitted scaling law to predict the optimal number of parameters to configure our released FLAME-MoE model family (§\ref{sec:family}).
In all following experiments, we use the data sampled from the state-of-the-art open-source pretraining corpus, DataComp-LM (DCLM)~\cite{li2024datacomp}, to train our models.

\subsection{IsoFLOP Profiles}
\label{sec:isoflop}

% \cx{we should plot our scaling law study curves? both methods used to explore scaling law can be plotted. also the downstream performances to confirm the generalization ability.} 

The first approach generates IsoFLOP profiles by varying the number of active parameters $N_{\text{active}}$ while keeping the computational budget $C$ fixed. 
For each fixed FLOPs budget, we calculate the number of training tokens $D$ as a function of $N_{\text{active}}$ using the formula:
\begin{equation}
    D = \frac{C}{\kappa N_{\text{active}}}
    \label{eq:compute_budget}
\end{equation}
where $\kappa$ is a constant commonly set to 6 in previous works~\cite{Chinchilla}.
By plotting the final validation loss achieved by each model configuration against its $N_{\text{active}}$ for a given FLOPs budget, we obtain an IsoFLOP curve. 
We then fit a parabolic function to each of these curves. 
The minimum of this parabola provides an estimate of the optimal $N_{\text{active}}$ for that specific computational budget. 
Finally, we fit a power law to these data points to model the relationship $N_{\text{active}}^*(C) \propto C^a$ and $D^*(C) \propto C^b$, 
where $N_{\text{active}}^*$ is the optimal $N_{\text{active}}$, $D^*$ is the optimal number of training tokens, $C$ is the computational budget, and $a, b$ are exponents to be determined.

In our experiments, we choose 4 computational budgets (1e18, 3e18, 6e18, and 3e19). For each budget, we train 16 models with different $N_{\text{active}}$, ranging from 33.4M to 1.7B.
The plotted IsoFLOP profiles are shown in Figure~\ref{fig:scaling-isoflop}, where the parabolas decently fit the data points.

\subsection{Parametric Loss Function Fitting}
\label{sec:parametric_loss}

The second approach fits a single parametric function to all experimental data points (final validation loss, $N_{\text{active}}$, number of training tokens) collected from the IsoFLOP experiments. 
Following~\cite{Chinchilla}, we model the final validation loss $\mathcal{L}_{\text{val}}$ as a function of $N_{\text{active}}$ and $D$:
\begin{equation}
     \mathcal{L}_{\text{val}}(N_{\text{active}}, D) = \frac{A}{(N_{\text{active}})^{\alpha}} + \frac{B}{D^{\beta}} + L_0
        \label{eq:parametric_loss}
\end{equation}
where $A, B, \alpha, \beta,$ and $L_0$ (an irreducible loss term) are parameters to be fitted from the experimental data.
To ensure the robustness of our fitted parametric loss function, 
we account for potential local minima by employing a grid search over initial parameter values for the fitting algorithm.

Furthermore, we utilize a Huber loss function with $\delta = 10^{-3}$ for the fitting. 
The Huber loss is less sensitive to outliers in the experimental data compared to a standard mean squared error loss, 
which is crucial for achieving generalized predictive performance when extrapolating to larger computational budgets.
As illustrated in Figure~\ref{fig:scaling-parametric}, the fitted parametric loss function precisely predicts loss values, demonstrating its effectiveness in capturing the underlying relationship between $N_{\text{active}}$, $D$, and $\mathcal{L}_{\text{val}}$.

Once this parametric loss function is fitted, we can use it to predict the optimal $N_{\text{active}}$ and training tokens $D$ for any given computational budget $C$. 
By minimizing $\mathcal{L}_{\text{val}}(N_{\text{active}},D)$ subject to the constraint $C = \kappa N_{\text{active}} D$, we can derive the optimal $N_{\text{active}}^*(C)$ and $D^*(C)$. Similar to the IsoFLOP approach, we can then fit power laws to these derived optimal values as a function of $C$.

\subsection{FLAME-MoE Model Family}
\label{sec:family}

We present our fitted scaling law in Figure~\ref{fig:scaling-fitted} and Table~\ref{tab:scaling}, 
where the results from both the IsoFLOP and parametric loss approaches closely align, demonstrating the robustness and consistency of our experimental methodology.
The generalization ability of our scaling law is also confirmed in Figure~\ref{fig:scaling-generalization}, 
where we observe a strong correlation (Spearman=0.89) between the validation loss and the performance on a representative downstream task, HellaSwag~\cite{zellers2019hellaswag}.

We choose three additional \#FLOPs from DCLM (2e19, 8e19, and 2.4e20) and apply our fitted scaling law to 
predict the optimal number of parameters under these \#FLOPs. 
The resulting compute-optimal models are FLAME-MoE-290M-1.3B, 721M-3.8B, and 1.7B-10.3B, where the first number denotes active parameters, and the second denotes total parameters.
% We scale the model's hidden size and the number of layers to create the FLAME-MoE family of models. 
% This allows us to vary model sizes for different research purposes while maintaining the architectural efficiency.
We then combine these three with the 4 compute-optimal models from our IsoFLOP experiments (FLAME-MoE-38M-100M, 98M-349M, 115M-459M, and 419M-2.2B) to form FLAME-MoE model family. 
The detailed configurations for each model scale within the FLAME-MoE family are presented in Table~\ref{tab:config}.

\begin{table}
    \caption{Performance of FLAME-MoE on downstream tasks across different compute budgets. 1: DCLM 400M-1x; 2: DCLM 400M-4x; 3: DCLM 1B-1x. The other 4 \#FLOPs are what we used in our scaling law experiments. We report accuracy for each task. Best performances are marked \textbf{bold}.
    % \cx{add non compute optimal MoE for more baselines} 
    % \cx{maybe call Dense-xx Dense-XX to make it clear that these are dense model we pretrained and evaluated ourselves. they are not DCLM released baselines.}
    }
    \label{tab:pretrain-main}
    \centering
    \resizebox{1.0\textwidth}{!}{%
      \begin{tabular}{l|l|ccccccc}
      \toprule
      \textbf{Model} & \textbf{\#FLOPs} & \textbf{ARC-E} & \textbf{ARC-C} & \textbf{OBQA} & \textbf{HellaSwag} & \textbf{PIQA} & \textbf{WinoGrande} & \textbf{Average} \\
      \midrule
      Dense-77M & 1.0e18 & 0.3266 & 0.2193 & \textbf{0.2460} & \textbf{0.2588} & 0.5582 & 0.4949 & 0.3506 \\
      FLAME-MoE-38M-100M & 1.0e18 & \textbf{0.3607} & \textbf{0.2235} & \textbf{0.2460} & 0.2582 & \textbf{0.5615} & \textbf{0.4949} & \textbf{0.3575} \\
      \midrule
      Dense-77M & 3.0e18 & 0.3754 & 0.2065 & 0.2440 & 0.2632 & 0.5832 & 0.5043 & 0.3628 \\
      FLAME-MoE-98M-349M & 3.0e18 & \textbf{0.4217} & \textbf{0.2150} & \textbf{0.2660} & \textbf{0.2790} & \textbf{0.6137} & \textbf{0.5241} & \textbf{0.3866} \\
      \midrule 
      Dense-191M & 6.0e18 & 0.4318 & 0.2099 & \textbf{0.2780} & 0.2769 & 0.6023 & 0.4917 & 0.3818 \\
      FLAME-MoE-115M-459M & 6.0e18 & \textbf{0.4651} & \textbf{0.2406} & 0.2700 & \textbf{0.3103} & \textbf{0.6240} & \textbf{0.5201} & \textbf{0.4050} \\
      % \midrule
      % \multicolumn{2}{l}{2.0e19 \#FLOPs}
      \midrule
      Dense-411M & 2.0e19$^1$ & 0.4802 & 0.2415 & 0.2940 & 0.3237 & 0.6398 & \textbf{0.5130} & 0.4154 \\
      FLAME-MoE-290M-1.3B & 2.0e19$^1$ & \textbf{0.5198} & \textbf{0.2474} & \textbf{0.3020} & \textbf{0.3567} & \textbf{0.6676} & 0.5036 & \textbf{0.4329} \\
      \midrule 
      Dense-411M & 3.0e19 & 0.5046 & 0.2568 & 0.2940 & 0.3353 & 0.6387 & \textbf{0.5099} & 0.4232 \\
      FLAME-MoE-419M-2.2B & 3.0e19 & \textbf{0.5476} & \textbf{0.2841} & \textbf{0.3100} & \textbf{0.3936} & \textbf{0.6806} & 0.5059 & \textbf{0.4536} \\
      % \midrule
      % \multicolumn{2}{l}{8.0e19 \#FLOPs}\\
      \midrule
      Dense-411M & 8.0e19$^2$ & 0.5484 & 0.2568 & 0.2940 & 0.3776 & 0.6654 & \textbf{0.5257} & 0.4447 \\
      FLAME-MoE-721M-3.8B & 8.0e19$^2$ & \textbf{0.5955} & \textbf{0.2884} & \textbf{0.3200} & \textbf{0.4536} & \textbf{0.6986} & 0.5130 & \textbf{0.4782} \\
      % \midrule
      % \multicolumn{2}{l}{2.4e20 \#FLOPs }\\
      \midrule
      Dense-1.4B & 2.4e20$^3$ & 0.6035 & 0.3020 & 0.3360 & 0.4653 & 0.6844 & 0.5178 & 0.4848 \\
      FLAME-MoE-1.7B-10.3B & 2.4e20$^3$ & \textbf{0.6469} & \textbf{0.3174} & \textbf{0.3500} & \textbf{0.5304} & \textbf{0.7236} & \textbf{0.5359} & \textbf{0.5174} \\
      \bottomrule
      \end{tabular}
    }
\end{table}
    \section{Pretraining FLAME-MoE}

In this section, we present the setup (§\ref{sec:exp-setup}), evaluation results (§\ref{sec:exp-results}), and throughput (§\ref{sec:throughput}) of our pretraining experiments.

\subsection{Experimental Setup}

\label{sec:exp-setup}
Our pretraining implementations are based on Megatron-LM~\cite{shoeybi2019megatron}, 
a highly optimized and integrated platform designed for large-scale training. 
We compare the performance of FLAME-MoE with dense baseline models of similar sizes, which are configured following the same architectures used in Pythia~\cite{biderman2023pythia} (Dense-77M and Dense-191M) or DCLM~\cite{li2024datacomp} (Dense-411M and Dense-1.4B). 
We train these baselines ourselves using the same codebase as FLAME-MoE to facilitate a fair comparison. 
For training, we use the Adam optimizer~\cite{KingBa15} with a maximum learning rate of 3e-4, a global batch size of 1024, and a sequence length of 2048.
The learning rate is configured using a WSD scheduler~\cite{hu2024minicpm}, with a warmup ratio of 0.01 and a decay ratio of 0.1 relative to the total number of training steps.
We use 32 NVIDIA H100 GPUs for training and store 10 checkpoints across evenly splitted training steps to study its performance trends. 
We provide a summary of all training configurations in Table~\ref{tab:training}.

% \cx{need more experimental details. What is LR? LR scheduling? optimizer? DO NOT missing implementation details in paper! List some here and use a full table in the appendix.}

For evaluation, we adapt lm-evaluation-harness~\cite{eval-harness} 
to work with models trained using Megatron-LM. 
We assess the performance of FLAME-MoE on 6 downstream tasks from OLMoE~\cite{muennighoff2024olmoe}\footnote{We replace MMLU~\cite{hendryckstest2021} with OBQA as the MMLU performance of small models is near random guess.}, 
including ARC-E~\cite{clark2018think}, ARC-C~\cite{clark2018think}, OBQA~\cite{OpenBookQA2018}, HellaSwag~\cite{zellers2019hellaswag}, PIQA~\cite{bisk2020piqa}, and WinoGrande~\cite{sakaguchi2021winogrande}. 
Following DCLM, we use 10-shot evaluation for ARC-E, ARC-C, HellaSwag and PIQA, and 0-shot for OBQA and WinoGrande.
The evaluation metric for all tasks is accuracy.

% \cx{should also add throughput and utilization measures of Megatron-MoE with different parallelization setups here?}

\begin{figure*}
    \centering
    \vspace{-0.2cm}
    \begin{subfigure}
        {0.32\textwidth}
        \centering
        \includegraphics[width=1.0\linewidth]{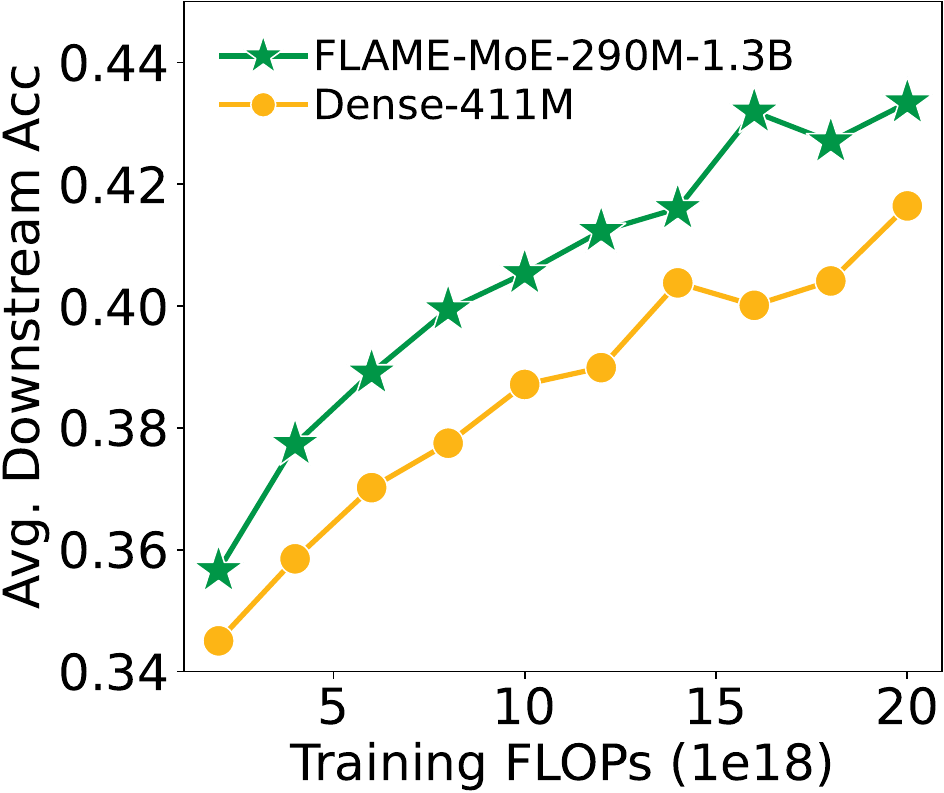}
        \caption{400M-1x (2.0e19 \#FLOPs).}
        \label{fig:400M_tokens}
    \end{subfigure}
    ~
    \begin{subfigure}
        {0.32\textwidth}
        \centering
        \includegraphics[width=1.0\linewidth]{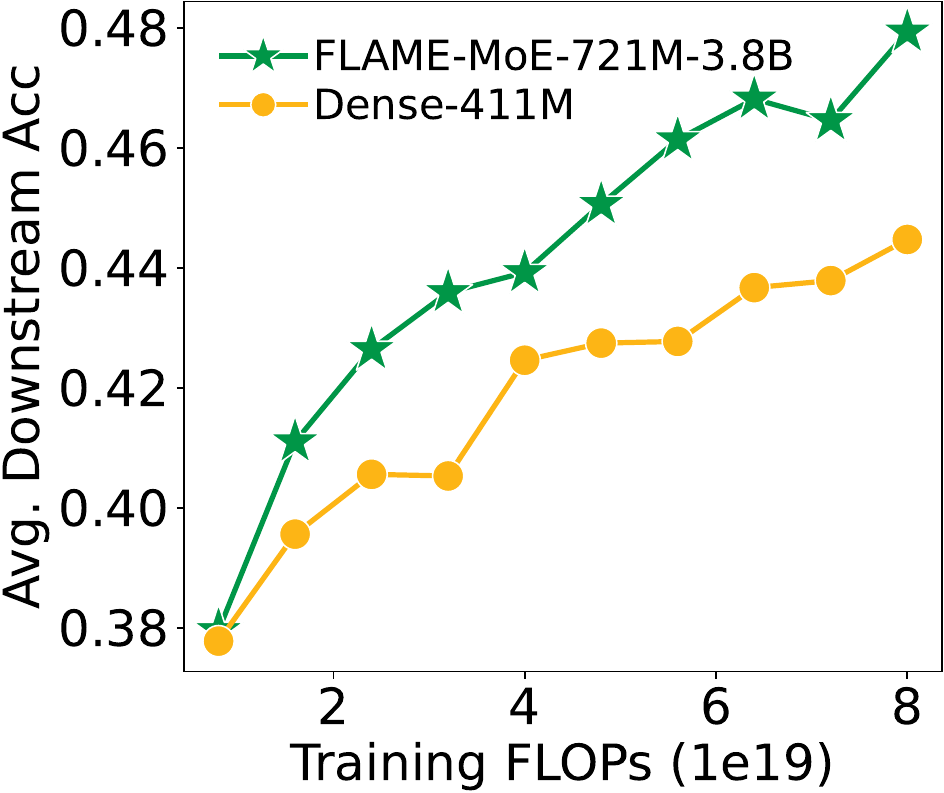}
        \caption{400M-4x (8.0e19 \#FLOPs).}
        \label{fig:1B_tokens}
    \end{subfigure}
    ~
    \begin{subfigure}
        {0.32\textwidth}
        \centering
        \includegraphics[width=1.0\linewidth]{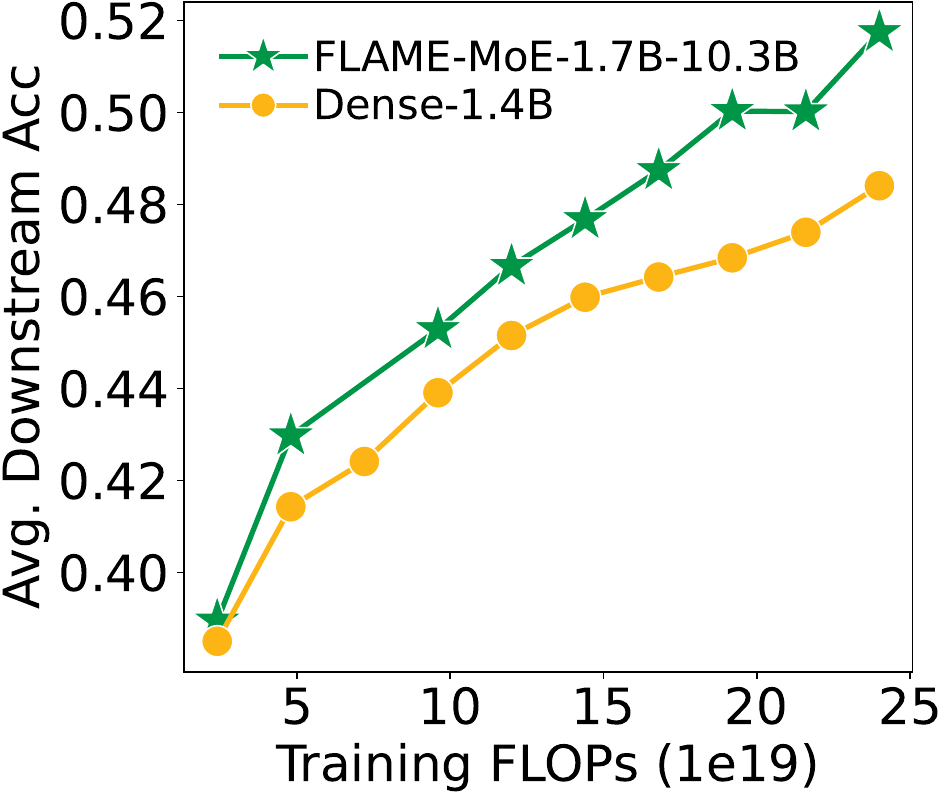}
        \caption{1B-1x (2.4e20 \#FLOPs).}
        \label{fig:400M_flops}
    \end{subfigure}
    \caption{Downstream comparison between FLAME-MoE and dense models during pretraining.}
    \label{fig:pretrain-main}
    % \vspace{-0.5cm}
\end{figure*}

\begin{figure*}
    \centering
    \begin{subfigure}{0.24\textwidth}
        \centering
        \includegraphics[width=\linewidth]{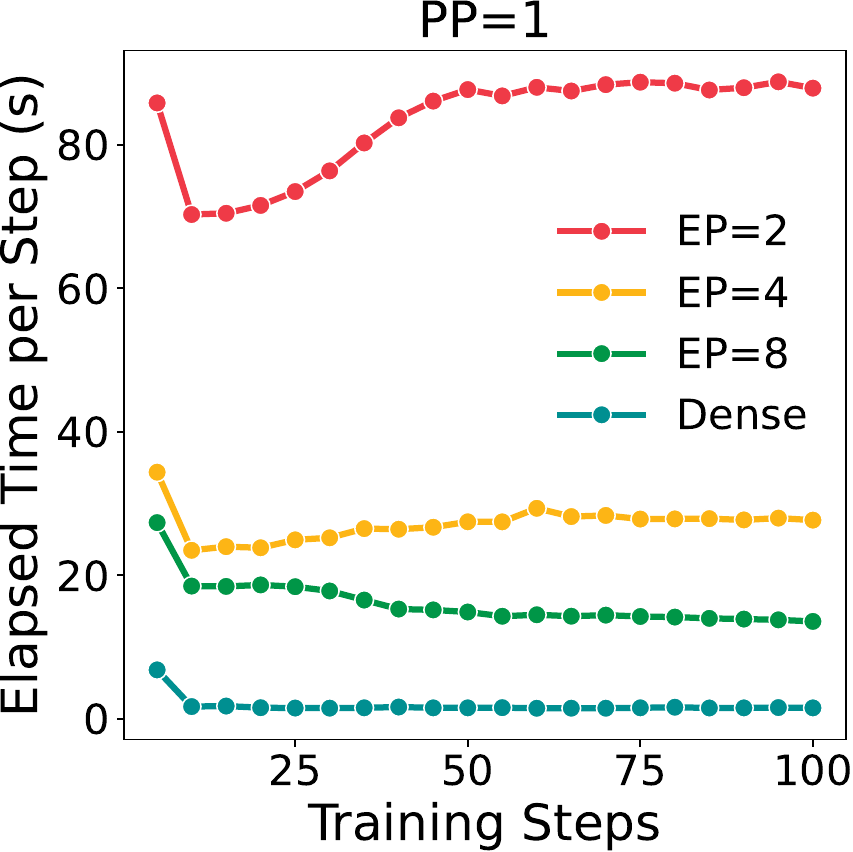}
    \end{subfigure}
    \hfill
    \begin{subfigure}{0.24\textwidth}
        \centering
        \includegraphics[width=\linewidth]{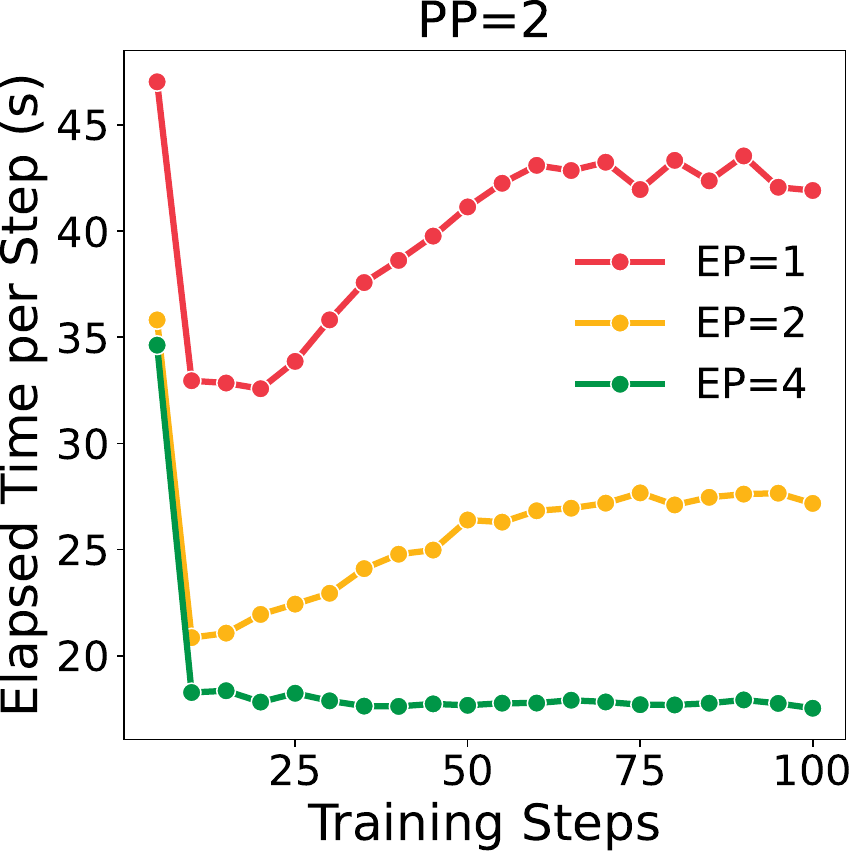}
    \end{subfigure}
    \hfill
    \begin{subfigure}{0.24\textwidth}
        \centering
        \includegraphics[width=\linewidth]{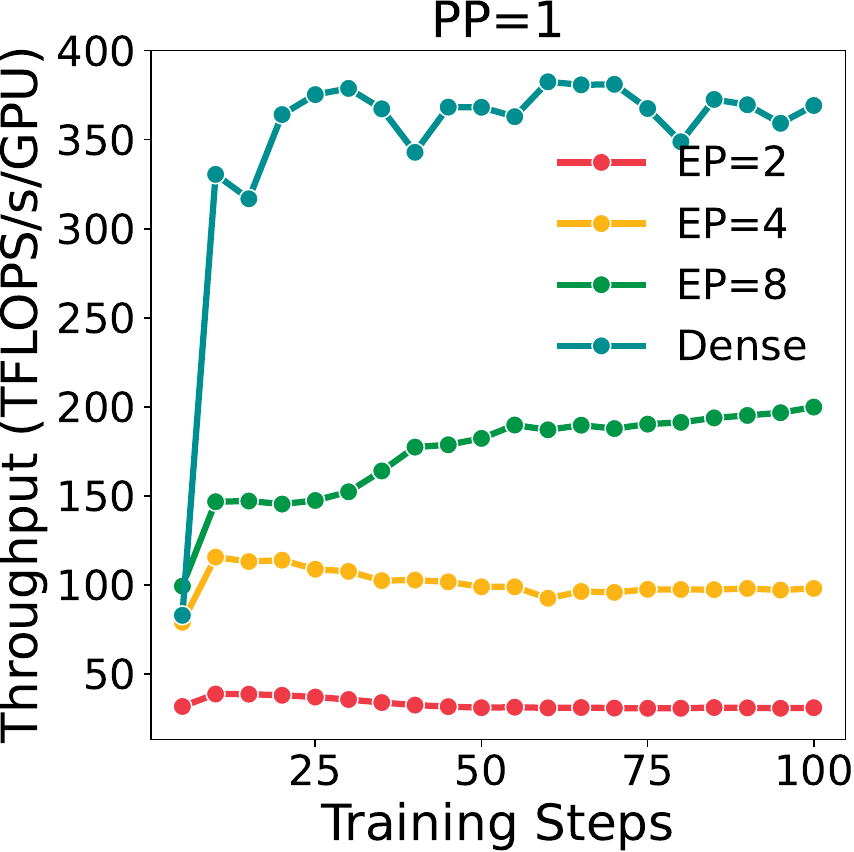}
    \end{subfigure}
    \hfill
    \begin{subfigure}{0.24\textwidth}
        \centering
        \includegraphics[width=\linewidth]{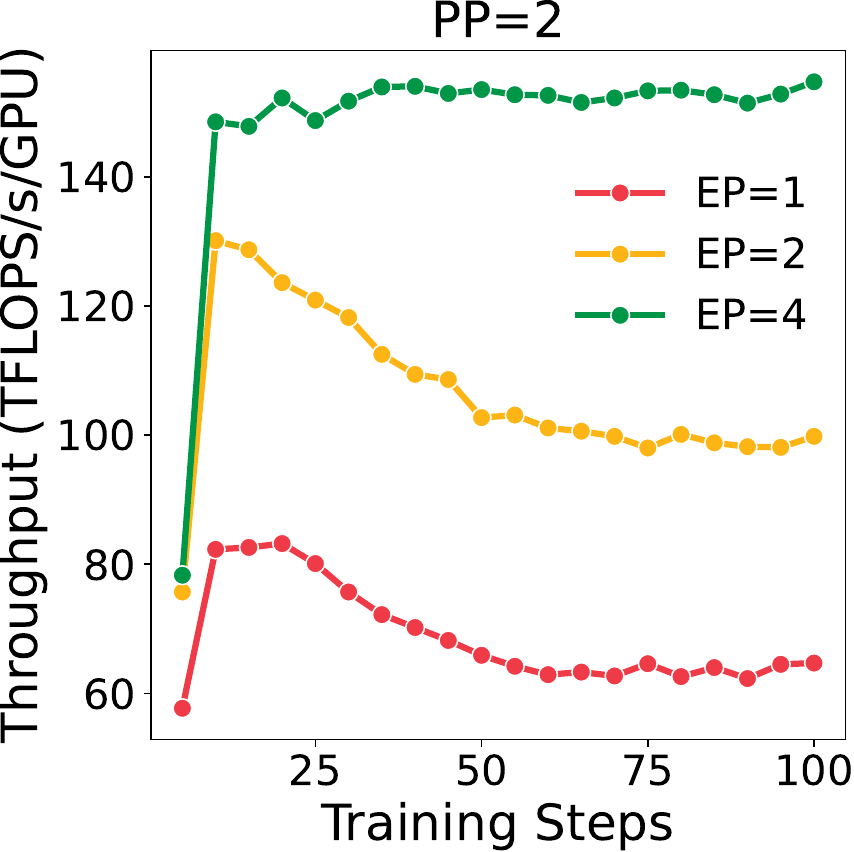}
    \end{subfigure}
    \caption{ Training efficiency of FLAME-MoE-1.7B-10.3B under different parallelization strategies (EP = Expert Parallel, PP = Pipeline Parallel). Dense-1.4B is also included here as a comparison.}
    \label{figure:utilization-comparison}
\end{figure*}

% \cx{add more details of experimental details. make it an individua lsection.}

\subsection{Evaluation Results}
\label{sec:exp-results}

We present the evaluation results of FLAME-MoE in Table~\ref{tab:pretrain-main}.
FLAME-MoE significantly outperforms the 
dense counterparts with the same pretraining FLOPs on almost every task, 
demonstrating the effectiveness of our scaling law and the superiority of MoE models.
The advantages of FLAME-MoE are more pronounced at larger scales, with more than 
3 points of average accuracy improvements over dense baselines under both 8.0e19 and 2.4e20 budgets.

We further illustrate the scaling curves of FLAME-MoE and dense models in Figure~\ref{fig:pretrain-main}. 
We observe that the performance gap between FLAME-MoE and dense models
continuously improves with the increase of pretraining FLOPs,
and FLAME-MoE can match or even outperform dense models trained with 2x FLOPs (e.g., in 400M-4x).
These results demonstrate that FLAME-MoE substantially improves pretraining efficiency, 
achieving a better speed-quality frontier.

\subsection{Training Throughput}
\label{sec:throughput}

In addition to task performance, we also evaluate the efficiency of training in various parallelization strategies. We profile both throughput (TFLOPs/s/GPU) and elapsed time per training step under different combinations of pipeline parallelism (PP) and expert parallelism (EP) that can fit into one node (8 GPUs). As shown in Figure~\ref{figure:utilization-comparison}, increasing EP generally improves utilization and reduces latency, while deeper pipeline parallelism (e.g., PP=2) can further enhance scalability. Based on these findings, we adopt the best-performing configuration of PP=1 and EP=8 for training our FLAME-MoE models, ensuring that our experiments efficiently utilize compute resources. 

However, while MoE models demonstrate great utilization under EP=8 as presented in Appendix~\ref{appendix:training-efficiency}, the overall FLOPs throughput still lags behind dense models. This discrepancy primarily arises from the inherent sparsity and communication overheads introduced by MoE architectures, which pose unique infrastructure challenges. These limitations highlight an area for improvement in open-source MoE implementations, such as those in Megatron-LM. Despite being among the most optimized open MoE frameworks available, current performance still trails that of proprietary systems with tightly integrated hardware-software co-design.

% \cx{we should mention that though utilization is high, the FLOPs throughput is not as good as dense models, due to the challenging infra from sparsity. We should discuss this honestly and clearly in a way that this is a place for improvements in open source MoE, that the current best Megaton-LM is still likely behind that of close source infrastructures.} \cx{this means we probablyu should include the dense model curves in Figure 3.} \cx{also do we have Utilization curves to show? we can show them in appendix and mention that it is high here in main.}

    \section{Empirical Analyses}

% A key advantage of the FLAME-MoE suite is that we release full pretraining checkpoints, which enables detailed empirical analysis of model behavior throughout training—not just at convergence. In this section, we leverage this capability to investigate how key MoE-specific phenomena, such as expert specialization (§\ref{sec:expert-specialization}), co-activation (§\ref{sec:expert-coactivation}), and router saturation (§\ref{sec:router-saturation}), evolve over the course of pretraining. Beyond these analyses, our released checkpoints offer a foundation for the broader community to study comprehensive dynamics in large-scale MoE models.

% One of the key contributions of the FLAME-MoE suite is the release of full pretraining checkpoints, which provides a rare opportunity to analyze model behavior throughout the entire training process—not just at convergence. In this section, we demonstrate how this resource enables in-depth investigations of MoE-specific dynamics, such as expert specialization (§\ref{sec:expert-specialization}), co-activation (§\ref{sec:expert-coactivation}), and router saturation (§\ref{sec:router-saturation}). These empirical analyses illustrate the kinds of insights made possible by FLAME-MoE and highlight its potential as a valuable tool for the broader research community studying large-scale MoE models.

A central advantage of the FLAME-MoE suite is the release of full pretraining checkpoints, which enables fine-grained analysis of model behavior across all stages of training—not just at convergence. In this section, we use this capability to explore key MoE-specific behaviors, including expert specialization (§\ref{sec:expert-specialization}), co-activation (§\ref{sec:expert-coactivation}), and router saturation (§\ref{sec:router-saturation}). These analyses exemplify how FLAME-MoE can support the broader academic community in studying the training dynamics of large-scale MoE models.

\subsection{Expert Specialization}
\label{sec:expert-specialization}

\begin{figure}
    \centering
    \includegraphics[width=\linewidth]{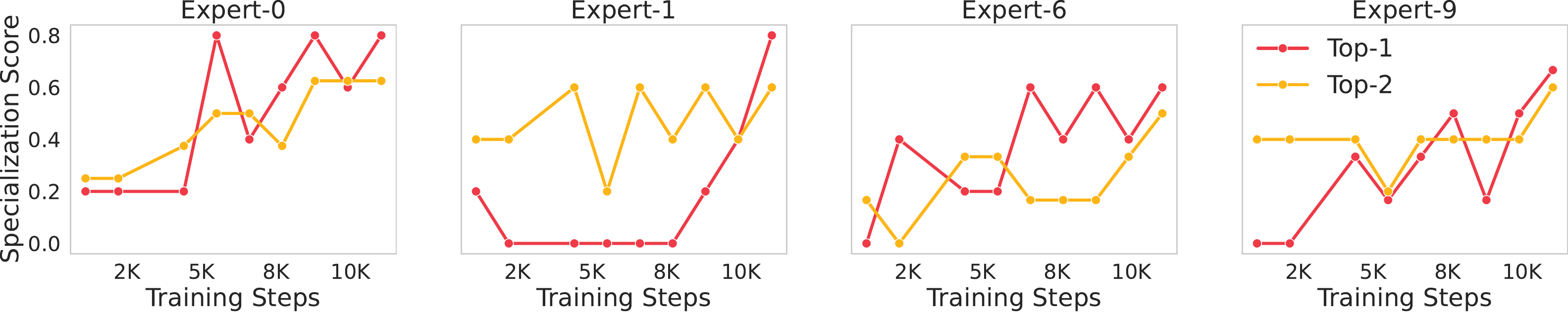}
    \caption{Evolution of specialization scores for the top-2 most specialized tokens across Experts 0, 1, 6, 9 at the final layer in FLAME-MoE-1.7B-10.3B on the validation set.}
    \label{figure:expert-specialization}
\end{figure}

To characterize expert behavior during pretraining, we analyze the routing patterns of individual tokens over time. In particular, we focus on whether certain experts consistently process specific tokens, which may indicate a form of specialization.

We define expert specialization as the fraction of times a given token is routed to a specific expert, normalized by the token's total frequency in the evaluation corpus. Formally, for a token $t$ and an expert $e$, the specialization score is:
\begin{equation}
    \text{Specialization}_e(t)=
    \frac{\text{\# times expert}~e~\text{activates on}~t}
    {\text{\# times}~t~\text{appears in the evaluation set}}
\end{equation}
This formulation quantifies how ``responsible'' an expert becomes for handling a particular token. The definition aligns closely with vocabulary specialization explored in OLMoE \cite{muennighoff2024olmoe}, but is evaluated here across full pretraining trajectories.

% \cx{this experiment is very unclear to me. both in description of what it is, and also justification of it.}

To track specialization over time, we fix the top-2 most specialized tokens for each expert at the end of pretraining and retrospectively evaluate their scores at earlier checkpoints. As shown in Figure~\ref{figure:expert-specialization}, we observe a consistent upward trend for these specialization scores across all experts analyzed. This indicates that token-level specialization gradually emerges and solidifies during pretraining. Additional details are presented in Appendix~\ref{appendix:expert-specialization}.

\subsection{Expert Co-activation}
\label{sec:expert-coactivation}

\begin{figure}
    \centering
    \includegraphics[width=\linewidth]{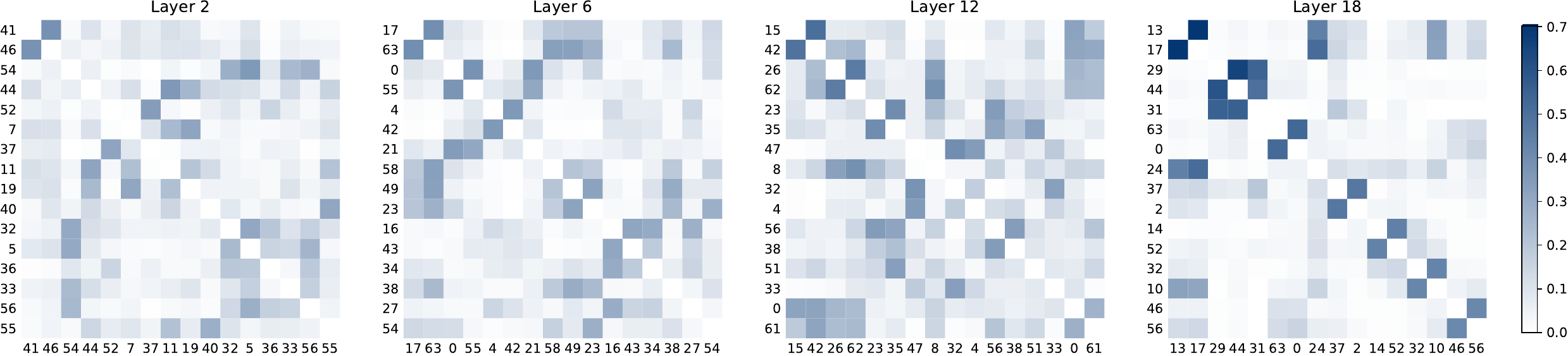}
    \caption{Expert co-activation in FLAME-MoE-1.7B-10.3B at the final checkpoint on the validation set. The heatmap shows pairwise co-activation scores among the 16 experts with the highest co-activation across layers 2, 6, 12, and 18. Expert IDs are shown on the axes.}
    \label{figure:expert-coactivation}
\end{figure}

% In top-$k$ routing schemes, each token is processed by multiple experts per layer. To better understand how experts interact within a layer, we study expert co-activation—the frequency with which pairs of experts are selected together for the same token. This metric reveals whether experts tend to act independently, redundantly, or cooperatively across the token distribution.

To understand expert interactions under top-$k$ routing, we analyze expert co-activation—how often expert pairs are selected together for the same token. This reveals whether experts behave independently or tend to co-operate. Following OLMoE~\cite{muennighoff2024olmoe}, we define the directional co-activation score from expert $E_i$ to expert $E_j$ as:
\begin{equation}
    \text{CoAct}(E_i, E_j)=
    \frac{|\{E_i \in \text{Top-}k(x),\, E_j \in \text{Top-}k(x) \mid x \in \mathcal{T}\}|}
    {|\{E_i \in \text{Top-}k(x) \mid x \in \mathcal{T}\}|}
\end{equation}
where $\mathcal{T}$ is the evaluation token set. 
CoAct measures the conditional likelihood that $E_j$ is co-activated with $E_i$. A high score indicates tight coupling between the two experts, whereas a low score suggests independence. 
Importantly, this formulation is asymmetric, i.e. $\text{CoAct}(i,j)\neq\text{CoAct}(j,i)$ in general.

As shown in Figure~\ref{figure:expert-coactivation}, co-activation is generally sparse, with most expert pairs exhibiting low scores. This suggests limited redundancy and indicates that experts are learning to be diverse instead of frequently overlapping in activation. In addition, co-activation increases with depth: the maximum score rises from 0.38 (Layer 2) and 0.39 (Layer 6) to 0.50 (Layer 12) and 0.70 (Layer 18). This pattern becomes more pronounced as training progresses; in Layer 18, the peak score grows from 0.51 to 0.70 between 10\% and 100\% of pretraining. Shallower layers show weaker trends. Additional results across different training steps and model scales are provided in Appendix~\ref{appendix:expert-coactivation}.

\subsection{Router Saturation}
\label{sec:router-saturation}

\begin{figure}
    \centering
    \includegraphics[width=\linewidth]{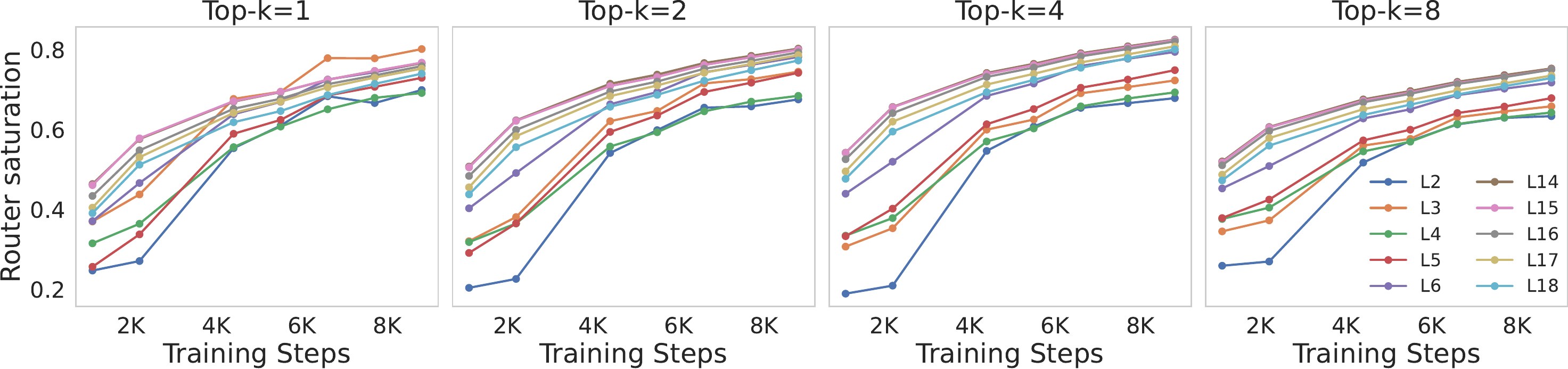}
    \caption{Router saturation across training for FLAME-MoE-1.7B-10.3B. Each subplot shows the average expert selection overlap with the final checkpoint using different top-$k$ values (1, 2, 4, 8).}
    \label{figure:router-saturation}
\end{figure}

Another key question in understanding MoE behavior is how early and how consistently the router converges on its expert selection patterns. Unlike expert co-activation, which focuses on the interactions among experts within a single forward pass, router saturation examines the temporal stability of routing decisions throughout pretraining. Understanding the saturation dynamics offers insight into how MoE models balance early specialization with continued learning capacity.

We define saturation as the average overlap between the top-$k$ experts selected for each token at step $t$ and at convergence. For a token $x$ and layer $l$, let $\text{Top-}k^{(t)}_l(x)$ denote the top-$k$ experts at step $t$, and $\text{Top-}k^{(T)}_l(x)$ the final selection. Then,

\begin{equation}
\text{Saturation}_l(t) =
\frac{1}{|\mathcal{T}|} \sum_{x \in \mathcal{T}} \frac{|\text{Top-}k^{(t)}_l(x) \cap \text{Top-}k^{(T)}_l(x)|}{k}
\end{equation}

Although FLAME-MoE was trained with top-$k=8$ routing, we report saturation under different evaluation settings ($k=1$, 2, 4, 8) to capture finer-grained changes in expert preference.

As shown in Figure~\ref{figure:router-saturation}, saturation increases steadily with training, with most layers reaching over 70\% agreement by the midpoint of training. Notably, saturation rises sharply within the first few thousand steps, suggesting that the router converges to stable expert assignments early in pretraining. This pattern is consistent across all top-$k$ settings, though absolute scores are higher for a smaller $k$, reflecting greater selection consistency among the most preferred experts. Deeper layers generally saturate faster than shallower ones, indicating more stable routing behavior as depth increases. Additional results across different model scales are provided in Appendix~\ref{appendix:router-saturation}.

% \cx{my concern is whether we have shown new insights versu OLMoE, as many of our analysis follwos there setting? we should emhasize on the insights one an derive from FLAME-MoE (maybe more about the pretraining dynamics and also comparison of obsevations across scales?) Is it possible to add cross scale resultss in our current analyses? that is sth OLMOE do not have for sure. If not, we should have them before we release on ArXiv.} 

    \section{Conclusion}

We present FLAME-MoE, a transparent and reproducible research platform built to advance the study of Mixture-of-Experts language models. By releasing a family of seven compute-optimal models along with complete training artifacts—including logs, checkpoints, routing traces, and evaluation scripts — we enable a rigorous, end-to-end study of MoE systems. Empirically, our models significantly outperform dense baselines under the same compute budgets, validating the effectiveness of our scaling law methodology that offers a principled guide for resource allocation. FLAME-MoE is not limited to isolated benchmark evaluation; rather, it offers an infrastructure to study every facet of MoE models, including expert specialization, routing dynamics, scaling behavior, training stability, parallel efficiency, and generalization performance. 
We believe that FLAME-MoE lays the groundwork for systematic, transparent exploration of sparse model architectures and invites broader community engagement in understanding and advancing MoE systems.

    \section*{Acknowledgements}
    We sincerely thank CMU Foundation and Language Model (FLAME) Center for providing support of computational resources. We sincerely thank Tianqi Chen, Ruihang Lai, Zihao Ye, Hongyi Jin, and Bohan Hou for discussing ideas and providing helpful feedback on this work. 
    
    \baselineskip 4.0mm
    \bibliographystyle{plainnat}
    \bibliography{bibliography}
    \appendix
\clearpage

\begin{table}[t]
    \centering
    \begin{minipage}{0.5\textwidth}
        \centering
        % \begin{table}
% \begin{minipage}{1.0\linewidth}
    \small
    \centering
    \caption{Scaling law parameters of FLAME-MoE.}
    \resizebox{0.6\textwidth}{!}{%
    \begin{tabular}{lr}
    \toprule
     Name & Value \\
    \midrule
     $A$ & 148.413257  \\
     $B$ & 3269017.372472  \\
     $L_0$ & 2.241716  \\
     $\alpha$ & 0.279702  \\
     $\beta$ & 0.715500  \\
     $a$ & 0.689902 \\
     $b$ & 0.310098 \\
    \bottomrule
    \end{tabular}
    }
    \label{tab:scaling}
% \end{minipage}
% \end{table}
    \end{minipage}%
    \hfill
    \begin{minipage}{0.5\textwidth}
        \centering
        % \begin{table}
% \begin{minipage}{1.0\linewidth}
    \small
    \centering
    \caption{Training details of FLAME-MoE.}
    % For all models, we set the total number of experts as 64 and activate 8 experts per token, where 2 of 8 are shared experts.
    \resizebox{0.7\textwidth}{!}{%
    \begin{tabular}{lr}
    \toprule
     Name & Value \\
    \midrule
     Batch Size & 1024  \\
     Sequence Length & 2048  \\
     Optimizer & Adam  \\
     Scheduler & WSD  \\
     Max Learning Rate & 3e-4  \\
     Min Learning Rate & 3e-5  \\
     Warmup Ratio & 0.01  \\
     Decay Ratio & 0.1  \\
     Pipeline Parallel & 1  \\
     Expert Parallel & 8  \\
     Device & 32xH100  \\
    \bottomrule
    \end{tabular}
    }
    \label{tab:training}
% \end{minipage}
% \end{table}
    \end{minipage}
\end{table}

\section{Training Efficiency} \label{appendix:training-efficiency}

\begin{figure}[h]
    \centering
    \begin{subfigure}{0.48\linewidth}
        \centering
        \includegraphics[width=\linewidth]{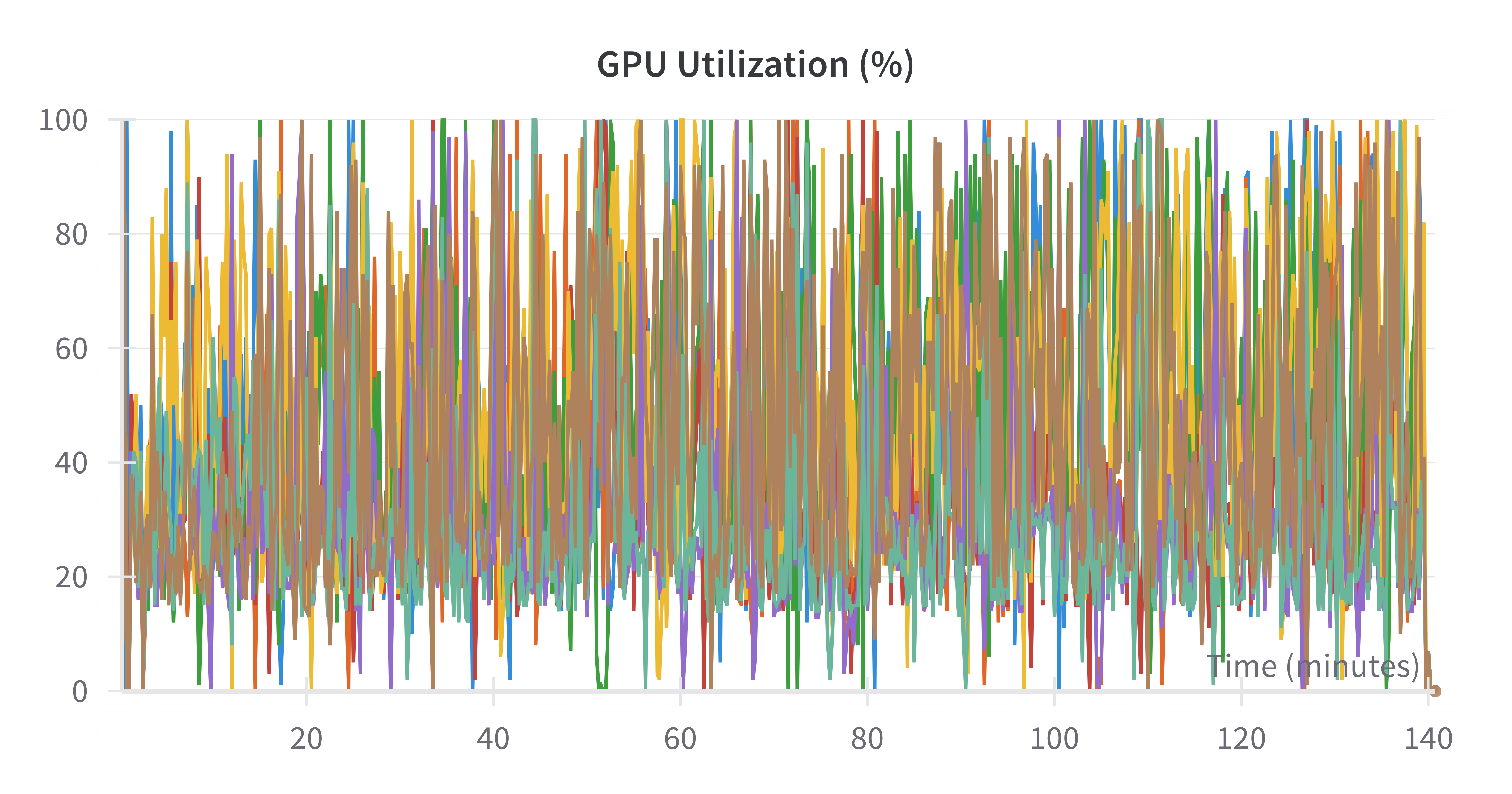}
        \caption{EP=2, PP=1}
    \end{subfigure}
    \hfill
    \begin{subfigure}{0.48\linewidth}
        \centering
        \includegraphics[width=\linewidth]{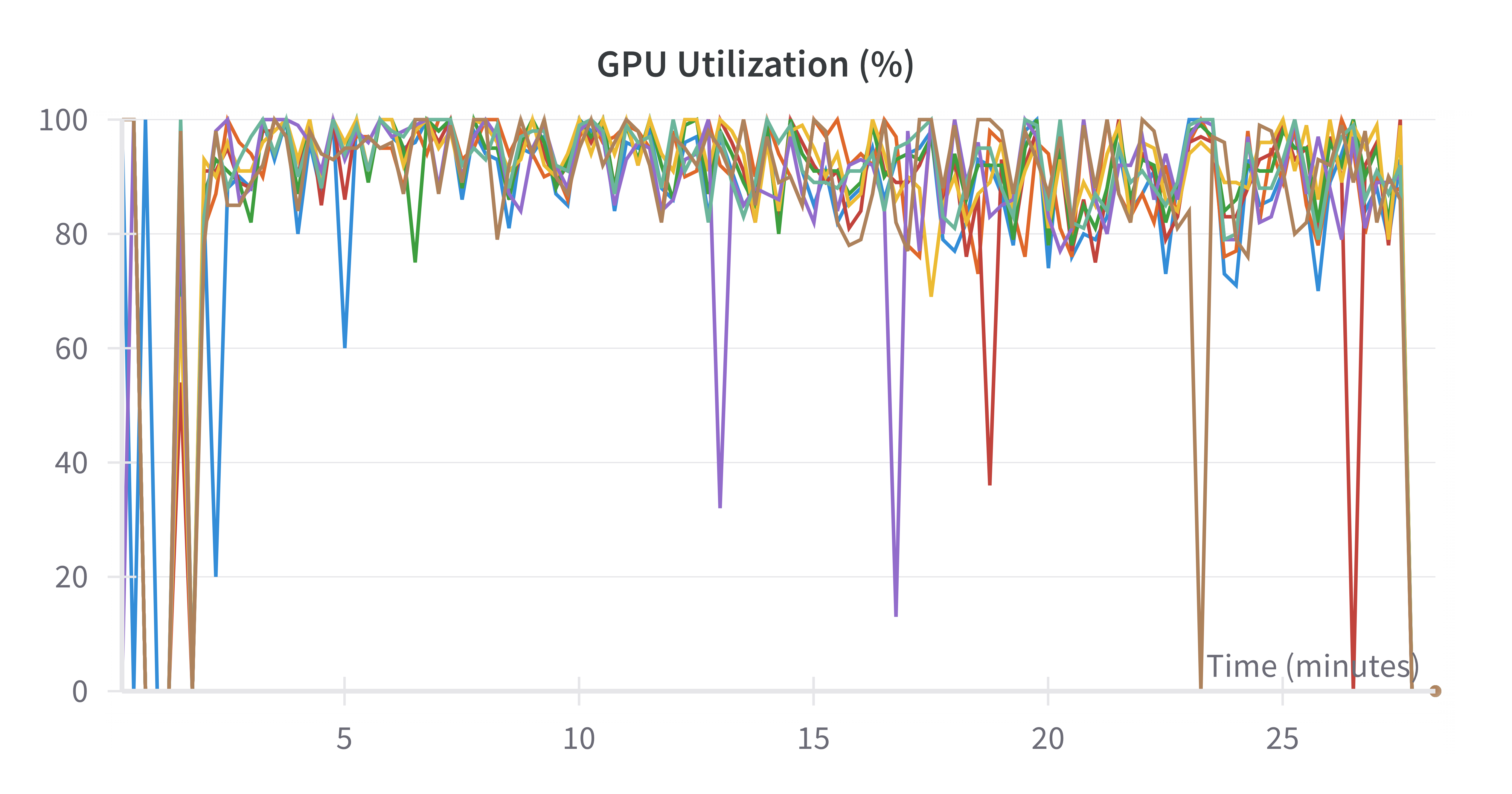}
        \caption{EP=8, PP=1}
    \end{subfigure}
    \caption{GPU utilization under different parallelization strategies (EP = Expert Parallel, PP = Pipeline Parallel) using Megatron-LM, collected by Wandb.}
    \label{figure:gpu-utilization}
\end{figure}

Figure~\ref{figure:gpu-utilization} illustrates the variation in GPU utilization under different levels of expert parallelism while keeping the pipeline parallelism constant. As observed, increasing the number of expert parallel groups (EP) from 2 to 8 leads to a marked improvement in GPU utilization. Specifically, in the configuration with EP=8 and PP=1, the majority of GPU utilization samples exceed 90\%, indicating consistently high utilization across devices.

\section{Expert Specialization} \label{appendix:expert-specialization}

\begin{figure}
    \centering
    \includegraphics[width=\linewidth]{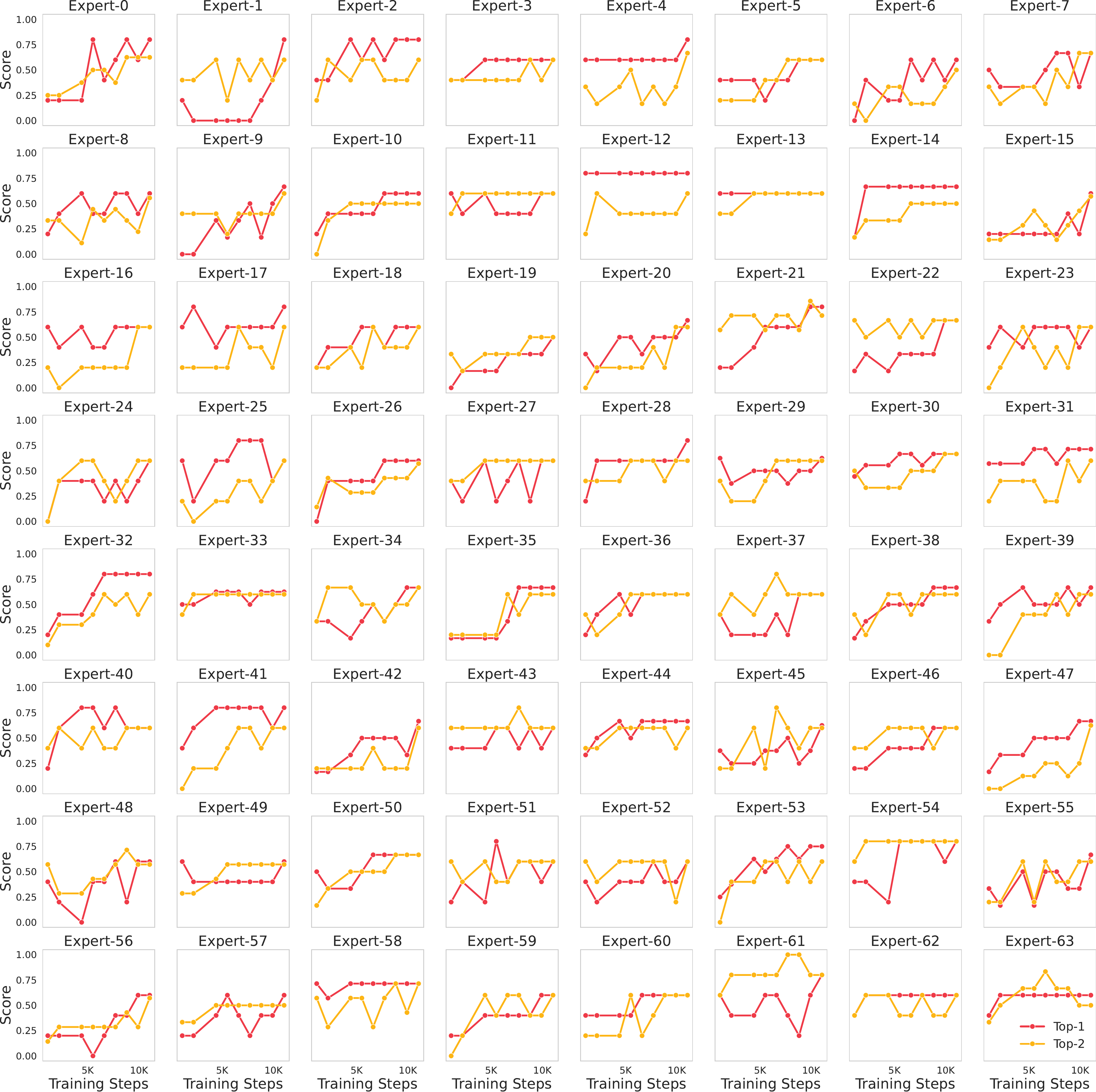}
    \caption{Evolution of expert specialization scores over the course of training in the final layer of FLAME-MoE-1.7B-10.3B. Each subplot tracks the specialization score of one expert across training steps, measured for its top-1 and top-2 most specialized tokens. The score reflects how consistently each expert is selected for specific tokens.}
    \label{figure:expert-specialization-everyone}
    % \vspace{-1cm}
\end{figure}

Figure~\ref{figure:expert-specialization-everyone} shows how expert specialization develops over training in the final layer of FLAME-MoE-1.7B-10.3B. Each plot represents an individual expert’s specialization score trajectory, computed for its top-1 and top-2 most frequently routed tokens. The results indicate that specialization progress is heterogeneous across experts—some develop clear specialization early, while others follow a more gradual or fluctuating trajectory. Nonetheless, a general upward trend is evident in the majority of experts, particularly for the top-1 and top-2 tokens. This suggests that, over time, most experts become more confidently and consistently associated with a subset of tokens, supporting the emergence of functional specialization within the MoE architecture.

\section{Expert Co-activation} \label{appendix:expert-coactivation}

\begin{figure}
    \centering
    \includegraphics[width=\linewidth]{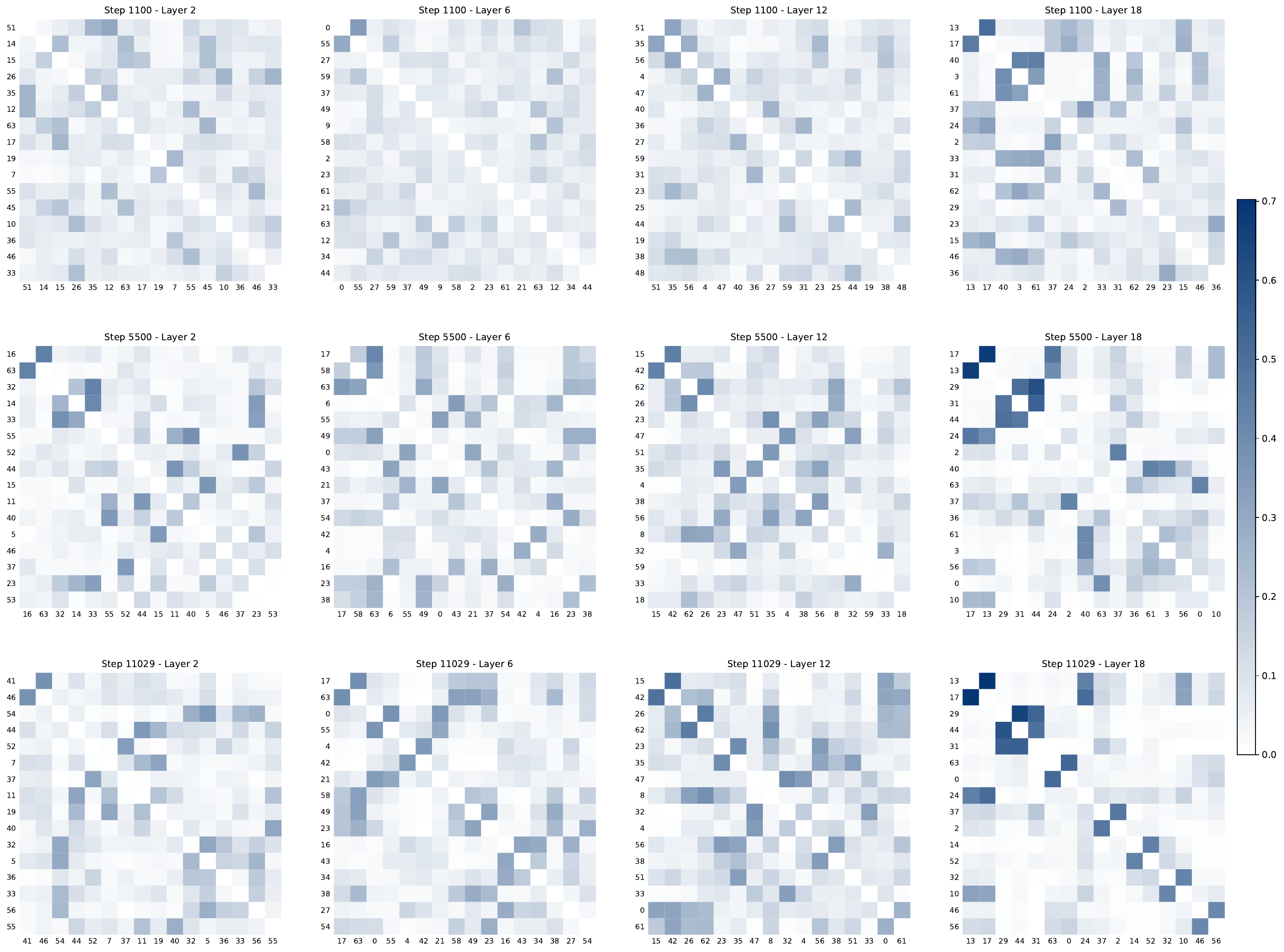}
    \caption{Evolution of expert co-activation in FLAME-MoE-1.7B-10.3B at 10\%, 50\%, and 100\% of pretraining (steps 1{,}100, 5{,}500, and 11,029), shown across layers 2, 6, 12, and 18. Each heatmap visualizes pairwise co-activation frequencies among the 16 most frequently co-activated experts at that layer and training stage.}
    \label{figure:expert-coactivation-overtime}
\end{figure}

Figure~\ref{figure:expert-coactivation-overtime} presents expert co-activation heatmaps at 10\%, 50\%, and 100\% of pretraining (corresponding to steps 1{,}100, 5{,}500; and 11{,}029, respectively). These visualizations illustrate the evolution of routing dynamics across training. As pretraining progresses, deeper layers display increasingly sharp co-activation patterns, with only a small subset of expert pairs being consistently co-activated. In contrast, co-activation in shallower layers remains more diffuse throughout. Notably, we do not observe global or widespread co-activation—most heatmaps are sparsely activated, indicating that the model does not route inputs uniformly across all expert pairs.

% \section{Cross-Scale Analysis of Expert Co-activation}
\begin{figure}[h]
    \centering
    \includegraphics[width=\linewidth]{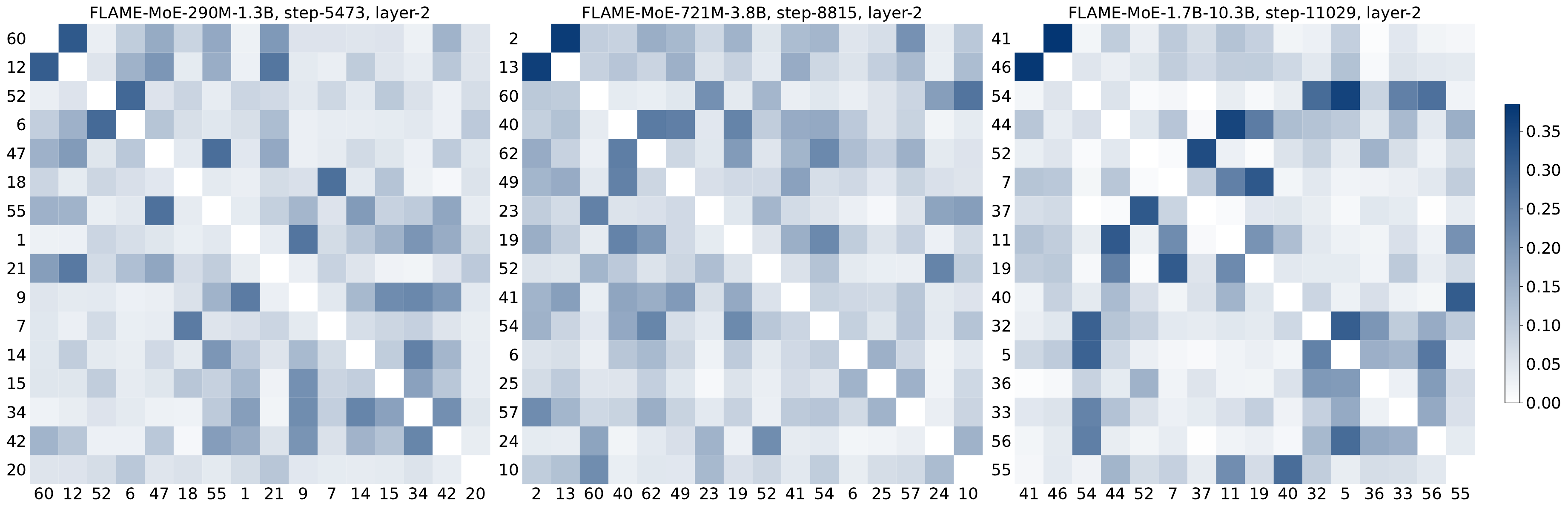}
    \caption{Expert co-activation patterns in FLAME-MoE models of varying scales at the final checkpoint on the validation set. Each heatmap shows pairwise co-activation scores among the 16 most frequently co-activated experts in the first MoE layer.}
    \label{figure:cross-scale-expert-coactivation-first-layer}
\end{figure}
\begin{figure}
    \centering
    \includegraphics[width=\linewidth]{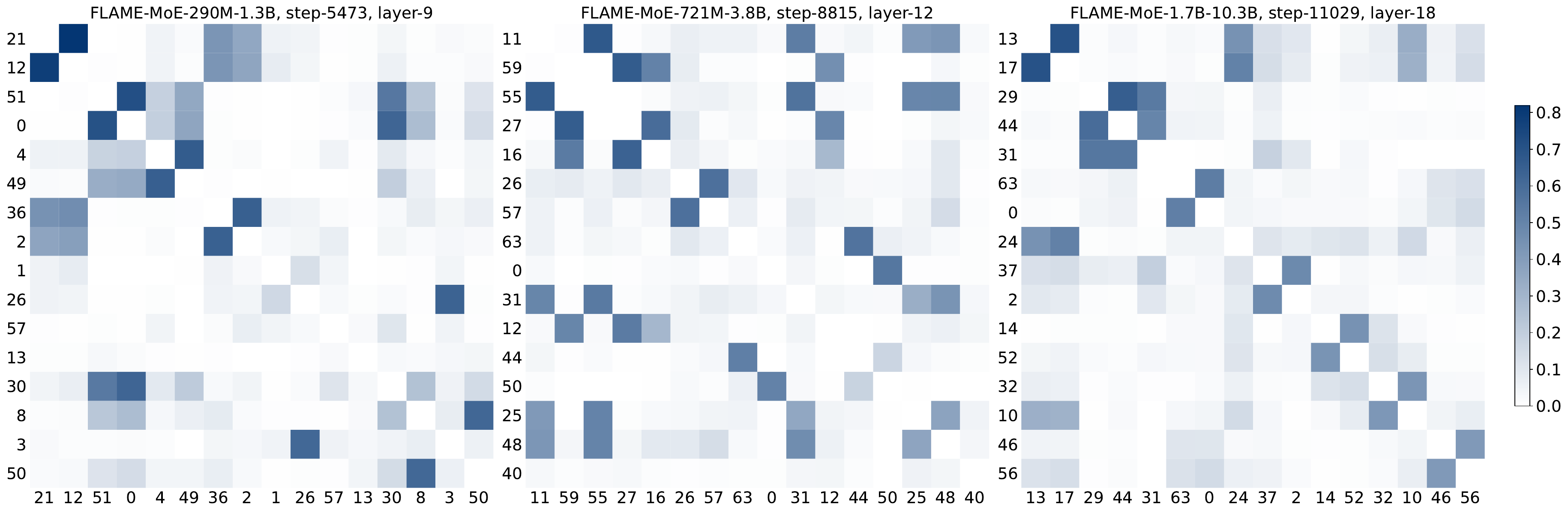}
    \caption{Expert co-activation patterns in FLAME-MoE models of varying scales at the final checkpoint on the validation set. Each heatmap shows pairwise co-activation scores among the 16 most frequently co-activated experts in the final MoE layer.}
    \label{figure:cross-scale-expert-coactivation-final-layer}
\end{figure}

Figures~\ref{figure:cross-scale-expert-coactivation-first-layer} and~\ref{figure:cross-scale-expert-coactivation-final-layer} illustrate expert co-activation patterns across FLAME-MoE models of different scales—FLAME-MoE-290M-1.3B, FLAME-MoE-721M-3.8B, and FLAME-MoE-1.7B-10.3B—at both shallow and deep layers. The visualizations present pairwise co-activation scores among the 16 most frequently co-activated experts within the first and final Mixture-of-Experts (MoE) layers for each model, respectively. Darker shades in the heatmaps denote stronger co-activation, with expert indices shown along both axes.

At shallow layers (e.g., layer 2), larger models exhibit broader and more intense expert co-activation, as seen in the FLAME-MoE-1.7B-10.3B model, which shows darker and more widespread patterns. This suggests that larger models tend to engage more experts early in processing, likely to capture more diverse or complex input features. In contrast, at deeper layers, the co-activation patterns appear relatively similar across model scales, with no pronounced differences, indicating a more consistent expert utilization behavior regardless of model size.

\section{Router Saturation}
\label{appendix:router-saturation}

\begin{figure}
    \centering
    \includegraphics[width=\linewidth]{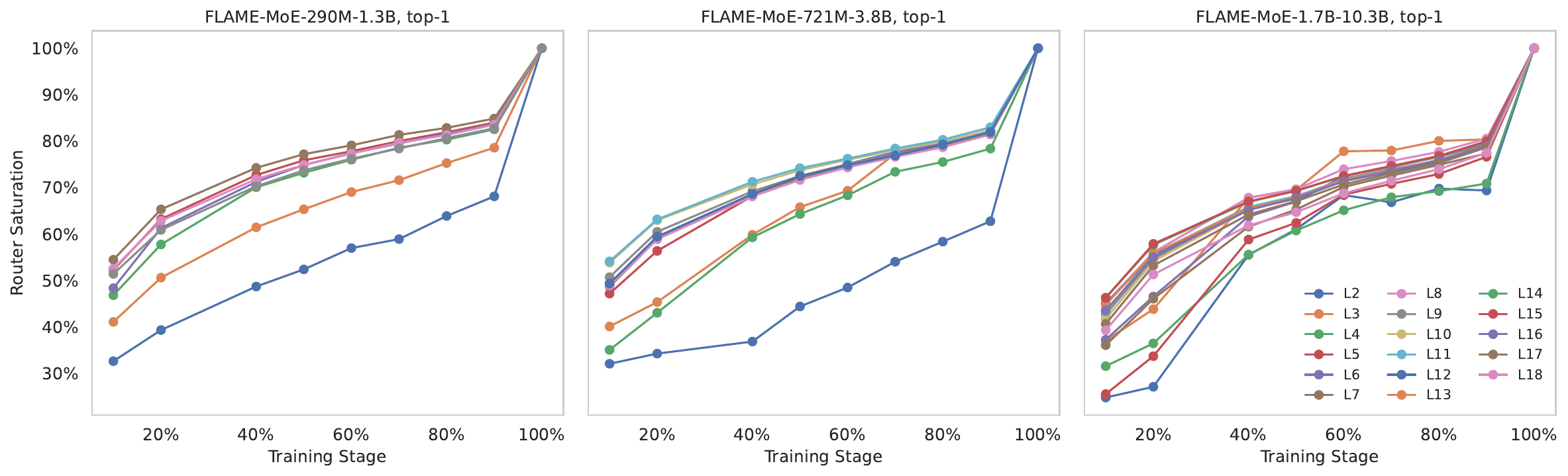}
    \caption{Router saturation over the course of training for FLAME-MoE models of different scales under top-1 routing. Each line represents a different MoE layer, showing the trend of average expert selection overlap with the final checkpoint.}
    \label{figure:cross-scale-router-saturation-top-1}
\end{figure}

\begin{figure}
    \centering
    \includegraphics[width=\linewidth]{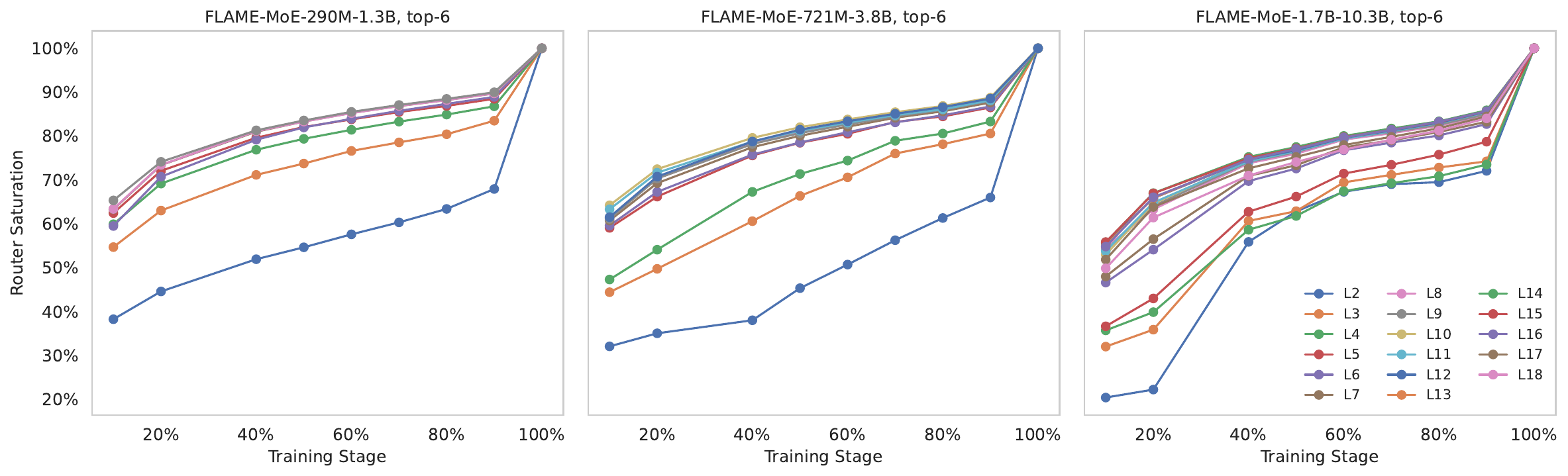}
    \caption{Router saturation over the course of training for FLAME-MoE models of different scales under top-6 routing. Each line represents a different MoE layer, showing the trend of average expert selection overlap with the final checkpoint.}
    \label{figure:cross-scale-router-saturation-top-6}
\end{figure}

Figures~\ref{figure:cross-scale-router-saturation-top-1} and~\ref{figure:cross-scale-router-saturation-top-6} show how router saturation evolves during training across FLAME-MoE models of different scales under top-1 and top-6 routing, respectively. Each line represents the saturation trend of a specific MoE layer, measured as the average overlap in expert selection with the final checkpoint.

We observe that larger models exhibit slower saturation, particularly in early layers. For example, in FLAME-MoE-1.7B-10.3B under top-1 routing, the saturation in layer 2 (blue line) starts notably lower compared to smaller models, indicating a greater degree of route plasticity early in training. More broadly, as model scale increases, a greater number of shallow layers show delayed saturation—evidenced by a wider set of lines starting at lower values—suggesting increased flexibility in routing decisions during early stages of training.

These trends imply that larger models maintain more potential for route refinement throughout training, especially in initial layers, which may contribute to their improved capacity for learning complex patterns and adapting expert utilization over time.

% \section{Societal Impacts} \label{appendix:societal-impacts}

% FLAME-MoE advances the societal impact of language model research by establishing a transparent, reproducible platform for studying Mixture-of-Experts architectures. Unlike prior systems constrained by proprietary components or limited release artifacts, FLAME-MoE provides comprehensive access to model checkpoints, training scripts, routing logs, and evaluation pipelines across multiple scales. This level of openness enables rigorous investigation into MoE-specific phenomena—such as expert specialization and routing dynamics—by researchers operating under diverse computational and institutional constraints. By lowering the barrier to entry for empirical MoE research and promoting verifiable findings, FLAME-MoE supports a more methodologically grounded and accessible trajectory for developing sparse, efficient language models.

    % \input{sections/checklist}
\end{document}